%% file: main.tex
\documentclass[journal]{IEEEtran}
\input{preamble}
\input{acronyms}

\newif\ifanonymized
\anonymizedfalse

\begin{document}

\title{Let the Dynamics Flow: \\ Stable Flow Matching Dynamical Systems}

\ifanonymized
\author{Author Names Omitted for Anonymous Review}
\else
\author{Rodrigo~P\'erez-Dattari$^{1}$,
        Francisco~Leiva$^{2}$,
        Andrea~Testa$^{3}$,
        Leonel~Rozo$^{4}$,
        Javier~Ruiz-del-Solar$^{2}$,
        No\'emie~Jaquier$^{1}$%
\thanks{$^{1}$Department of Robotics, Perception, and Learning, KTH Royal Institute of
Technology, Stockholm, Sweden. {\tt\small \{rpd, jaquier\}@kth.se}}
\thanks{$^{2}$Advanced Mining Technology Center (AMTC) and Department of Electrical Engineering, Universidad de Chile, Santiago, Chile. {\tt\small \{francisco.leiva, jruizd\}@ing.uchile.cl}}
\thanks{$^{3}$Bosch Center for Artificial Intelligence, Renningen, Germany. {\tt\small andrea.testa@de.bosch.com}}
\thanks{$^{4}$The Italian Institute of Artificial Intelligence (AI4I), Turin, Italy. {\tt\small leonel.rozo@ai4i.it}}
\thanks{This work was partially supported by the Wallenberg AI, Autonomous Systems and Software Program (WASP) funded by the Knut and Alice Wallenberg Foundation. The computations were enabled by the Berzelius resource provided by the Knut and Alice Wallenberg Foundation at the National Supercomputer Centre. F.L. and J.R.-d.S. were supported by FONDECYT project 1251823, and ANID-PIA project CIA250010.}
}
\fi

\markboth{}%
{}

\maketitle

\begin{abstract}
Flow matching has recently emerged as a powerful approach for imitation learning, enabling scalable, expressive, and multimodal motion policies. However, when modeling these policies as dynamical systems, incorporating formal stability guarantees into these generative models is a prerequisite to ensure safe and generalizable robot behaviors, which remains a significant challenge. 
This paper introduces Stable Flow Matching Dynamical Systems (SFMDS), a novel framework that bridges the gap between highly expressive generative modeling and formal stability guarantees. SFMDS parametrizes dynamical systems via flow matching while constraining the model to satisfy positive invariance and/or Lyapunov stability conditions. We propose two variants: a soft constraint based on a penalty term, and a hard structural constraint embedded directly into the model architecture. We further extend both formulations to Lie groups to robustly handle orientation trajectories. Experiments on benchmark datasets, in simulation, and on a humanoid robot show that SFMDS learns stable, scalable, and multimodal dynamical systems in low- and high-dimensional state spaces, enabling safe and expressive robot motion generation. SFMDS matches or outperforms state-of-the-art methods on unimodal datasets, while substantially improving performance on multimodal datasets, where competing approaches fail to capture multi-modal behaviors. 
Accompanying source code and video are available at:
\href{https://let-the-dynamics-flow.github.io/SFMDS/}{\textcolor{blue}{https://let-the-dynamics-flow.github.io/SFMDS/}}.
\end{abstract}


\IEEEpeerreviewmaketitle

\input{Sections/01_Introduction}
\input{Sections/02_Background}
\input{Sections/03_Problem_Formulation}
\input{Sections/04_Method_imitation_learning}

\input{Sections/05_Method_A}
\input{Sections/06_Method_B}
\input{Sections/07_Method_C}
\input{Sections/08_Experiments}
\input{Sections/10_Conclusions}

\input{Sections/09_Appendix/Appendix}

\bibliographystyle{plainnat}
\bibliography{references}

\end{document}

%% file: preamble.tex
\input{math_commands.tex}

\usepackage{url}
\usepackage{times}
\usepackage{overpic}

\usepackage[numbers]{natbib}
\usepackage{multicol}
\usepackage{balance}
\usepackage[bookmarks=true]{hyperref}

\usepackage{multicol}
\usepackage{amssymb}
\usepackage{bm}
\usepackage{amsmath}
\usepackage{amsthm}
\usepackage{graphicx}
\usepackage[dvipsnames]{xcolor}
\usepackage{import}
\usepackage{mathtools}
\usepackage{svg}
\usepackage{soul}
\usepackage{makecell}
\usepackage{multirow}
\usepackage{array}
\usepackage{colortbl}
\usepackage{siunitx}
\usepackage{adjustbox}
\usepackage{adjustbox}
\usepackage{booktabs}
\usepackage[nolist]{acronym} 
\usepackage{tikz}
\usepackage[ruled, vlined]{algorithm2e}
\usepackage[caption=false]{subfig}
\usepackage{mdframed}
\usepackage[most]{tcolorbox}
\definecolor{lightlightsteelblue}{rgb}{0.88,0.92,0.97}
\definecolor{verylightsteelblue}{rgb}{0.94,0.96,0.99}
\definecolor{conditionframe}{rgb}{0.45,0.55,0.70}

\tcbset{
  remarkbox/.style={
    enhanced, breakable,
    colback=gray!5, colframe=gray!50,
    boxrule=0pt, leftrule=2pt, arc=2pt,
    left=8pt, right=8pt, top=6pt, bottom=6pt,
  },
  conditionbox/.style={
    enhanced, breakable,
    colback=verylightsteelblue, colframe=blue!15!gray!35!white,
    boxrule=0pt, leftrule=2pt, arc=2pt,
    left=8pt, right=8pt, top=6pt, bottom=6pt,
  }
}

\newtheorem{condition-inner}{Condition}

\newenvironment{condition}
  {\begin{tcolorbox}[conditionbox]\begin{condition-inner}}
  {\end{condition-inner}\end{tcolorbox}}
\usetikzlibrary{arrows.meta}

\newcommand{\trsp}{\mathsf{T}}  

\newcommand{\euclideanspace}{\mathbb{R}}
\newcommand{\torus}{\mathbb{T}}

\newcommand{\admfuncset}{\mathcal{F}}

\usepackage[colorinlistoftodos]{todonotes}

\newtheoremstyle{boldremark}
  {}{}                  
  {\normalfont}          
  {}                    
  {\bfseries}            
  {.}                   
  { }                   
  {}                    

\theoremstyle{boldremark}
\newtheorem{remark}{Remark}

\def\he{{\bm{h}_\tau}}
\def\dhe{{\dot{\bm{z}}_\tau}}
\def\hL{{\tilde{\bm{h}}_\tau}}
\def\uL{{\tilde{u}_\theta}}
\def\dX{{\dot{\mathcal{X}}}}
\def\dXL{\dot{\mathcal{L}}}
\def\dXaonly{{\dot{\mathcal{X}}_{\textnormal{A}}}}
\def\dXa{{\dot{\mathcal{X}}_{\textnormal{A}}(\bm{x}_t; \theta)}}
\def\dXaL{{\dot{\mathcal{L}}_{\textnormal{A}}(\bm{y}_t; \theta)}}
\def\dXaLonly{{\dot{\mathcal{L}}_{\textnormal{A}}}}
\def\dXaLB{{\dot{\mathcal{L}}^{\sB}_{\textnormal{A}}}}
\def\dUa{{\mathcal{U}}_{\textnormal{A}}(\bm{h}_{\tau}, \bm{x}_t ; \theta)}
\def\dUaonly{{\mathcal{U}}_{\textnormal{A}}}
\def\dUaL{{\tilde{\mathcal{U}}}^{\sB}_{\textnormal{A}}(\tilde{\bm{h}}_\tau)}

\def\dXan{{\dot{\mathcal{X}}_{\textnormal{A}}}}
\def\dUan{{\mathcal{U}}_{\textnormal{A}}}

\def\xe{\bm{x}_{\mathnormal{e}}}

\def\ye{\bm{y}_{\mathnormal{e}}}

\definecolor{grayblue}{HTML}{8BA4E1}
\DeclareRobustCommand{\grayrectangle}{\tikz{ \filldraw[color=gray!60, fill=gray!60, thick](0.05,0.06) rectangle (.3,0.2);}}
\DeclareRobustCommand{\graybluerectangle}{\tikz{ \filldraw[color=grayblue!60, fill=grayblue!60, thick](0.05,0.06) rectangle (.3,0.2);}}

%% file: math_commands.tex
\usepackage{amsmath,amsfonts,bm}

\def\1{\bm{1}}

\DeclareMathAlphabet{\mathsfit}{\encodingdefault}{\sfdefault}{m}{sl}
\SetMathAlphabet{\mathsfit}{bold}{\encodingdefault}{\sfdefault}{bx}{n}

\def\sA{{\mathcal{A}}}
\def\sB{{\mathcal{B}}}

\def\sL{{\mathcal{L}}}

\def\sN{{\mathcal{N}}}

\def\sS{{\mathcal{S}}}
\def\sT{{\mathcal{T}}}
\def\sU{{\mathcal{U}}}

\def\sX{{\mathcal{X}}}

\def\sZ{{\mathcal{Z}}}

\DeclareMathOperator*{\argmin}{arg\,min}

%% file: acronyms.tex
\newacro{fm}[FM]{flow matching}
\newacro{sfm}[SFM]{stable flow matching}
\newacro{sfmds}[SFMDS]{Stable Flow Matching Dynamical Systems}
\newacro{ode}[ODE]{ordinary differential equation}
\newacro{node}[NODE]{Neural ODE}
\newacro{cnf}[CNF]{Continuous Normalizing Flows}

%% file: Sections/01_Introduction.tex
\section{Introduction}

Flow matching~\citep{Lipman23:FlowMatching, liuflow2023} has recently emerged as a powerful generative model in downstream tasks due to its scalability, expressivity, and simple training procedure. In robotics, flow matching policies have demonstrated exceptional performance in synthesizing robot motion skills via imitation learning, efficiently learning complex multimodal action distributions and outperforming classical behavior cloning approaches~\citep{braun2024riemannian, chisari2024learning, funk2024actionflow,  yan2025maniflow}. Despite its strengths, flow matching is not inherently designed for robot motion generation and control and thus lacks rigorous safety guarantees necessary when modeling robot policies.
However, previous works on robot motion learning recognized the relevance of introducing stability guarantees for controlled generalization by representing motion policies as learned dynamical systems~\citep{khansari-zadeh2011learning,mohammadi2023neural,perez2023stable,perrin2016fast,Rana2020:EuclideanizingFlows,Urain20:ImitationFlow}, thus ensuring safe, robust and generalizable robot behaviors~\citep{billard2022learning}. Yet, existing dynamical system-based models are often unimodal and feature limited expressivity compared to modern generative models. 
This paper bridges this gap by introducing flow matching dynamical systems that are stable by design.

\begin{figure}[t]
    \centering
    \includegraphics[width=\linewidth,trim=30 50 110 30,clip]{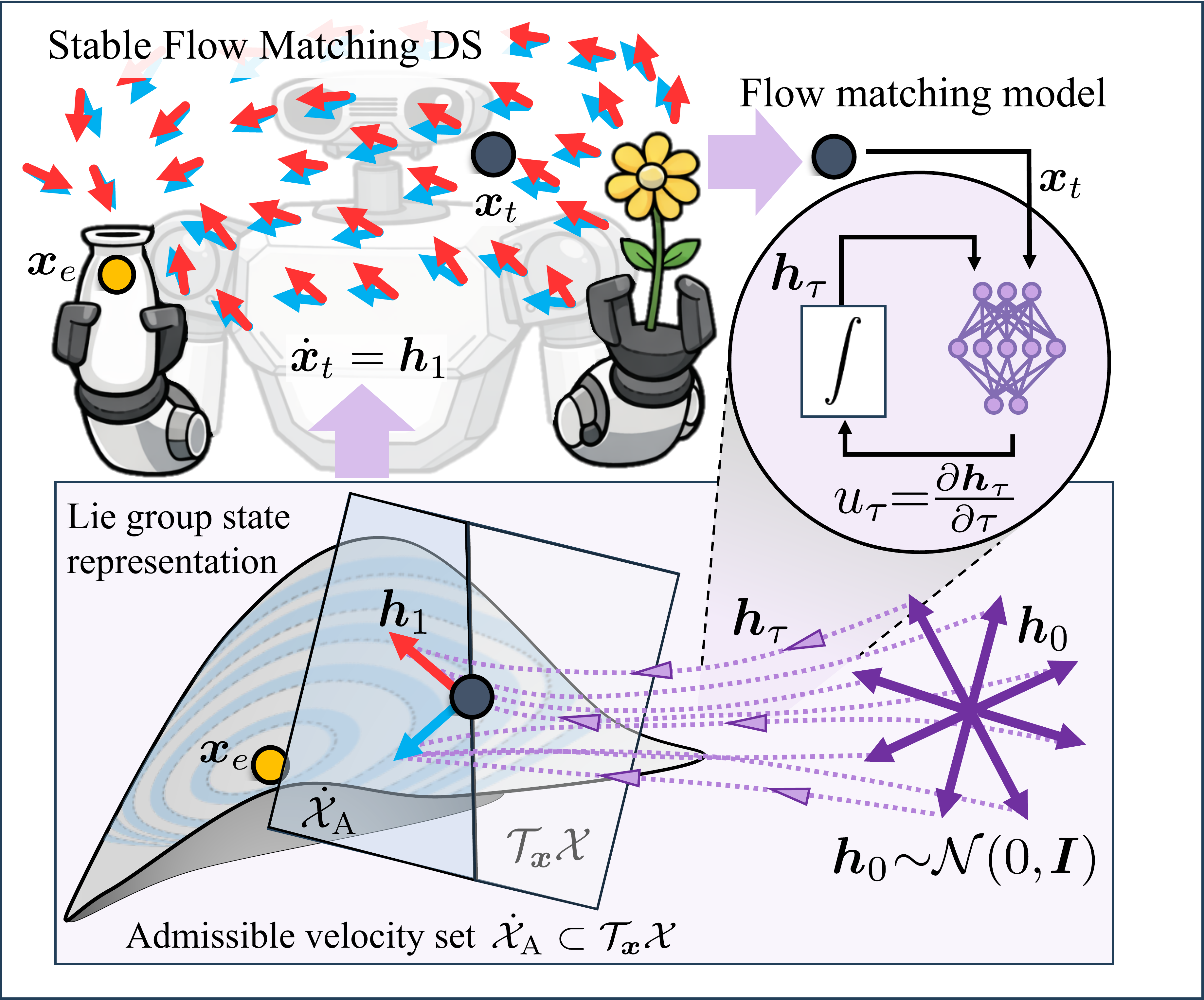}
    \caption{Robot behavior modeled as a stable flow matching dynamical system (SFMDS) on a Lie group. Velocities $\dot{\bm{x}}_t=\bm{h}_1$ are learned via flow matching. Asymptotic stability is enforced by constraining the solution space of flow matching to a set of admissible velocities $\dot{\sX}_{\text{A}}$ derived from a latent Lyapunov function. SFMDS generates multimodal behaviors, represented by blue and red arrows.}
    \label{fig:cover_figure}
\end{figure}

\textbf{Related work.} \citet{khansari-zadeh2011learning} pioneered time-invariant stable dynamical system learning by enforcing structural Lyapunov constraints on the parameters of systems modeled via Gaussian mixture models. To overcome the limited accuracy of this model, several approaches learned Lyapunov functions to shape or correct the target dynamics~\citep{khansari-zadeh2014learning, lemme2014neural, duan2017fast, kolter2019learning}. Alternatively, another line of works leveraged diffeomorphisms to transfer the stability of a simple system onto the target demonstrated dynamics~\citep{perrin2016fast, Rana2020:EuclideanizingFlows, Urain20:ImitationFlow,zhi2022diffeomorphic}. In this case, the model accuracy largely depends on the expressivity of the diffeomorphism, typically an invertible neural network. To allow for greater architectural flexibility, some methods opt for weak stability guarantees via regularization, at the expense of stable-by-design guarantees~\citep{perez2023stable, perez2024puma}. In this paper, we build on the foregoing approaches and enforce stability either weakly via regularization or strictly via diffeomorphisms.

While recent works learned dynamical systems via neural networks~\citep{perez2023stable, perez2024puma,Rana2020:EuclideanizingFlows, Urain20:ImitationFlow,zhang2022learning,zhi2022diffeomorphic}, they did not fully leverage the power of generative models. Several approaches notably borrow diffeomorphic backbones from normalizing flows~\citep{Papamakarios21:NF} for stability mapping~\citep{Rana2020:EuclideanizingFlows,Urain20:ImitationFlow,Urain22:StableSE3,zhang2022learning}. However, their dynamics remain either deterministic or driven by low-variance unimodal noise, thus falling short to encode complex, potentially multimodal policies.
In addition, normalizing flows training generally relies on explicit maximum-likelihood estimation that requires computing expensive inverse Jacobian determinants.
Flow matching instead simplifies training via a simulation-free objective, leading to state-of-the-art results~\citep{Lipman23:FlowMatching, liuflow2023}.

Recently, S$^2$Diff~\citep{Cheng25:s2diff} provided weak stability guarantees on diffusion models by guiding them via learned certificates. Unlike S$^2$Diff, which optimizes control policies with respect to a given cost function, our method learns stable policies from demonstrations. Moreover, we exploit the inherent simplicity of flow matching, avoiding the need to model the stochastic differential equations associated with diffusion models, while introducing both weak and strict stability formulations. Closer to our work are stable Riemannian flow matching policies~\citep{ding2025fast}, which equip the flow matching vector field with stability to the support of the target behavior distribution using LaSalle's invariance principle~\citep{sprague2024stable}. However, this method offers no stability guarantees for the underlying robot motion dynamics. Instead, inspired by~\citep{ding2025fast,sprague2024stable}, we employ LaSalle's principle to constrain the solution space of flow matching models. This enables the enforcement of stability guarantees for robot trajectories while capturing complex multimodal action distributions.

Robotics applications commonly require learned policies to control both end-effector position and orientation, which intrinsically belong to non-Euclidean spaces such as the Lie group $\mathrm{SE}(3)$. Yet, most approaches assume Euclidean state spaces, thereby rendering the stability guarantees futile and thus limiting their practical use. Recent works underscore the importance of geometric formulations for stable dynamical systems, representing states on Lie groups~\citep{Urain22:StableSE3,mohammadi2023neural} or on Riemannian manifolds~\citep{zhang2022learning}. Building on this, we design stable flow matching dynamical systems on Lie groups.

\textbf{Contributions.} This paper introduces \textit{\ac{sfmds}}, a novel generative model that combines the scalability and expressivity of flow matching with formal stability guarantees to learn stable, generalizable, and expressive multimodal dynamical systems. Unlike existing approaches that rely on explicit dynamics parameterizations, \ac{sfmds} implicitly models the dynamics via flow matching (Sec.~\ref{sec:method_a}), while constraining the learned solutions to an admissible set, defined by criteria from positive invariance and (global) asymptotic stability analysis (Sec.~\ref{sec:enforcing_inductive_biases}). Inspired by previous works~\citep{Rana2020:EuclideanizingFlows, Urain22:StableSE3, perez2024puma}, \ac{sfmds} enforces stability by integrating latent Lyapunov functions into flow matching (Sec.~\ref{sec:method-stability-enforcement}), and extends these concepts to dynamics evolving on Lie groups (Sec.~\ref{sec:sfmds_lie}). 
We test \ac{sfmds} on multiple benchmarks, and learn realistic motion skills featuring complex full-pose trajectories on $\mathrm{SE}(3)$, both in simulation and with a real humanoid robot. Finally, we demonstrate \ac{sfmds} scalability on a $1682$-dimensional state space reaction–diffusion system (Sec.~\ref{sec:experiments}).

\begin{mdframed}[hidealllines=true,backgroundcolor=lightlightsteelblue,innerleftmargin=.1cm,innerrightmargin=.1cm,innertopmargin=.1cm,innerbottommargin=0.1cm,, roundcorner=2pt]
In summary, the \textbf{main contributions} of this paper are: \emph{(1)} Soft \ac{sfmds} that weakly enforces stability via regularization; \emph{(2)} Hard \ac{sfmds} that guarantees stability via structural constraints; \emph{(3)} The extension of both soft and hard \ac{sfmds} to Lie groups, enabling stable learning of position and orientation dynamics; and \emph{(4)} Two new benchmark datasets: A multimodal dataset in $\euclideanspace^2$ and a Lie-group dataset in $\torus^2$. 
\end{mdframed}

%% file: Sections/02_Background.tex
\section{Background}
\label{sec:background}
\subsection{Learning Stable Dynamical Systems}
\label{subsec:dynamics-object}

We aim to design expressive, scalable, and stable dynamical systems to learn goal-driven robot motions. We represent time-invariant autonomous dynamical systems via an \ac{ode} defined by a learnable vector field $f_{\theta}$,\looseness-1
\begin{equation}
\label{eq:dynamics-as-model}
    \dot{\bm{x}}_t = \frac{d \bm{x}_t}{d t}=f_{\theta}(\bm{x}_t),
\end{equation}
where $\bm{x}_t \in \mathcal{X} \subseteq \euclideanspace^n$ denotes the state at time $t$, ${\dot{\bm{x}}_t \in \dot{\mathcal{X}} \subseteq \euclideanspace^n}$ is its time derivative, and $\theta$ are the parameters of the vector field $f_{\theta} \!:\! \mathcal{X} \!\to\! \dot{\mathcal{X}}$.
In this paper, we model the dynamics $f_{\theta}$ \emph{implicitly} via \textbf{flow matching} and focus on imposing two stability properties, introduced next.
\subsubsection{Positive Invariance}

A set $\sA \subseteq \sX$ is said to be \emph{positively invariant} w.r.t. $f_{\theta}$ if any trajectory initialized in $\sA$ and evolving according to the dynamical system~\eqref{eq:dynamics-as-model} always remains in $\sA$, i.e., ${\bm{x}_0 \in \sA \Rightarrow \bm{x}_t \in \sA, \forall\, t \ge 0}$. Positive invariance provides a set-based stability criterion for characterizing safety and boundedness of dynamical systems. 

\subsubsection{Asymptotic Stability}

Dynamical systems displaying asymptotic stability converge to an equilibrium $\xe\in\mathcal{X}$ for any initial state in $\mathcal{X}$ as $t \to \infty$. They can be characterized via Lyapunov stability analysis.
Formally, the equilibrium $\bm{x}_e$ is asymptotically stable for the dynamical system~\eqref{eq:dynamics-as-model} if there exists a continuously-differentiable Lyapunov function $V:\mathcal{X}\to\euclideanspace_{+}$\footnote{$\euclideanspace_{+}$ denotes the set of nonnegative real numbers.} such that
{\setlength{\abovedisplayskip}{6pt}
 \setlength{\belowdisplayskip}{6pt}
\begin{align}
\label{eq:lyapunov_1}
&V(\bm{x}_t)>0, \; \forall\bm{x}_t\neq \bm{x}_e, \quad V(\bm{x}_e)=0, \;\text{ and } \\
\label{eq:lyapunov_2}
&\dot{V}(\bm{x}_t)=\nabla_{\bm{x}} V(\bm{x}_t)^\trsp f_{\theta}<0, \; \forall\bm{x}_t\neq \bm{x}_e,  \quad \dot{V}(\bm{x}_e)=0.
\end{align}
}

\subsection{Flow Matching}
\label{subsec:flow-matching}
Flow matching~\citep{Lipman23:FlowMatching} is a simulation-free approach to train \ac{cnf}~\cite{kobyzev2020normalizing}, which  improves both training practicality and generative quality. As a \ac{cnf}, flow matching reshapes a simple base distribution $p_0$ into an arbitrarily complex and potentially multimodal target data distribution $p_1 = p_{\text{data}}$. This transformation is represented by a parameterized smooth bijection with a smooth inverse,  i.e., a \emph{diffeomorphism}, $\phi_\theta$. This diffeomorphism induces a \emph{pushforward} distribution $[\phi_\theta]_* p_0$, which describes the distribution obtained by applying $\phi_\theta$ to samples from $p_0$, based on the change-of-variables formula
\begin{equation}
    p_1 = [\phi_\theta]_* p_0, \qquad p_1(\bm{h})=p_0\!\left(\phi_\theta^{-1}(\bm{h})\right) \left| \det \bm{J}_{\phi_\theta^{-1}}(\bm{h}) \right|,
\end{equation}
where $\bm{J}_{\phi_\theta^{-1}}(\bm{h})$ denotes the Jacobian matrix of $\phi_\theta^{-1}$ with respect to $\bm{h}$.

In \ac{cnf}, the flow $\phi_\theta$ is constructed via a vector field ${u_\theta:\mathbb{R}^n \times [0,1]\to\mathbb{R}^n}$ as the solution of the \ac{ode}
\begin{equation}
\label{eq:IVP}
    \frac{d \bm{h}_\tau}{d \tau} = u(\bm{h}_\tau, \tau), \qquad \bm{h}_0 \sim p_0,
\end{equation}
where $\tau \in [0,1]$ is an auxiliary time variable, and $\bm{h}_\tau$ denotes the state along the flow at time $\tau$. The transformed sample is
then obtained as $\bm{h}_1=\phi_\theta(\bm{h}_0, \tau\!\!=\!\!1)$, where the base distribution $p_0$ is chosen to be simple and easy to sample from (e.g., a Gaussian distribution).

Given a desired \emph{probability path} $\{p_\tau\}_{\tau\in[0,1]}$, which interpolates between $p_0$ and $p_1$, and its associated vector field $u$, flow matching regresses a parametrized vector field $u_\theta$ to the target vector field $u$, which can be achieved by minimizing
\begin{equation}
\label{eq:FM_loss}
    \ell_{\text{FM}} = \mathbb{E}_{\tau \sim \text{Unif}(0,1),\, \bm{h}_\tau \sim p_\tau} \left[ \left\| u_\theta(\bm{h}_\tau,\tau) - u(\bm{h}_\tau,\tau) \right\|^2 \right].
\end{equation}
However, the objective~\eqref{eq:FM_loss} is intractable, since neither $p_\tau$ nor $u$ are available in closed form. Flow matching circumvents this issue by replacing the probability path with a family of \emph{conditional paths} $p_\tau(\bm{h}_\tau | \bm{h}_1)$, conditioned on the endpoint samples $\bm{h}_1 \sim p_1$, generated by a tractable conditional target vector field $u(\bm{h}_\tau, \tau; \bm{h}_1)$. This leads to the conditional flow matching loss
\begin{equation*}
\label{eq:CFM_loss}
    \ell_{\text{CFM}} = \mathbb{E}_{\tau \sim \text{Unif}(0,1),\, \bm{h}_\tau \sim p_\tau} \left[ \left\| u_\theta(\bm{h}_\tau,\tau) - u(\bm{h}_\tau,\tau;\bm{h}_1) \right\|^2 \right].
\end{equation*}
Regressing $u_\theta$ onto the conditional target yields the same gradient as the original objective~\eqref{eq:FM_loss}, while making the problem tractable~\citep{Lipman23:FlowMatching}.

In this work, we adopt a variant of rectified flows~\citep{liuflow2023}, in which the conditional path is further conditioned on a base sample $\bm{h}_0 \sim p_0$, so that the intermediate state follows the deterministic linear interpolation
\begin{equation}
\label{eq:gausspath}
    \bm{h}_\tau = (1-\tau)\bm{h}_0 + \tau \bm{h}_1.
\end{equation}
Differentiating with respect to $\tau$ gives the constant-velocity conditional target
\begin{equation}
\label{eq:gausstargetvf}
    u(\bm{h}_0,\bm{h}_1) = \bm{h}_1 - \bm{h}_0,
\end{equation}
leading to the conditional flow matching loss
\begin{equation}
\label{eq:CFM_loss}
    \ell_{\text{CFM}} = \mathbb{E}_{\substack{\tau \sim \text{Unif}(0,1) \\ \bm{h}_0 \sim p_0,\,\bm{h}_1 \sim p_1}} \left[ \left\| u_\theta(\bm{h}_\tau, \tau) - (\bm{h}_1 - \bm{h}_0) \right\|^2 \right].
\end{equation}
Finally, conditioning the vector field on an external variable $\bm{x}$ (e.g., a robot's state) yields a distinct field $u_{\theta}(\bm{h}_\tau, \tau; \bm{x})$ for each $\bm{x}$, intuitively enabling a separate generative process to be trained per $\bm{x}$.

\subsection{Stable Autonomous Flow Matching}
As we will describe in Sec.~\ref{sec:method_a}, we implicitly model the dynamics $f_{\theta}$~\eqref{eq:dynamics-as-model} via flow matching through the auxiliary vector field $u_{\theta}$. Consequently, constraining $f_{\theta}$ to satisfy the stability properties of Sec.~\ref{subsec:dynamics-object} reduces to constraining $u_{\theta}$. To this end, we build on~\citet{sprague2024stable}, who proposed a stable autonomous flow matching variant based on \emph{LaSalle's invariance principle}~\citep{LaSalle66}, to stabilize the flow to the support of the target distribution $p_1$. Although this variant was shown to enhance the robustness of robot flow-matching policies~\cite{ding2025fast}, it does not directly translate to the stability properties required for stable dynamical systems $f_{\theta}$ (Sec.~\ref{subsec:dynamics-object}). In Sec.~\ref{sec:enforcing_inductive_biases}, we instead apply LaSalle's principle to constrain $u_{\theta}$ to converge to admissible sets aligned with the stability properties required on $f_{\theta}$. Next, we proceed to describe LaSalle's invariance principle and stable autonomous vector fields as introduced in~\cite{sprague2024stable}.

\subsubsection{LaSalle's Invariance Principle}
\label{sec:lasalle}
Given a time-invariant vector field $u_{\theta}$, LaSalle's invariance principle~\citep{LaSalle66} states that if a continuously differentiable function ${H(\bm{h}_\tau):\euclideanspace^n\to\euclideanspace_{+}}$ satisfies the conditions
\begin{equation}
\label{eq:invariance_dxae}
\begin{aligned}
    \nabla_{\bm{h}} H(\bm{h}_\tau)^\trsp u_{\theta}(\bm{h}_\tau) < 0, \quad \forall \bm{h}_\tau \notin \sA,\\
    \nabla_{\bm{h}} H(\bm{h}_\tau)^\trsp u_{\theta}(\bm{h}_\tau) = 0, \quad \forall \bm{h}_\tau \in \sA,
\end{aligned}
\end{equation}
then, as $\tau \to \infty$, $\bm{h}_\tau \to \sA \subseteq \euclideanspace^n$, i.e., the flow converges to the set $\sA$.

\subsubsection{Stable Autonomous Vector Fields}
\label{sec:stable_vector_fields}
LaSalle's invariance principle assumes time-invariant vector fields; however, flow matching typically employs time-varying systems. Hence, the function $u_\theta$ is reparametrized as a time-invariant vector field by augmenting its state space $\bm{h}_\tau$ with a pseudo-time variable $s_\tau \in [0,1]$. This variable replaces the explicit (auxiliary) time input $\tau$ and is constrained to increase monotonically with $\tau$, thereby encoding the progression of time. The augmented state is denoted as $\bm{z}_\tau \!=\! [\bm{h}_\tau, s_\tau] \in \sZ$ with $\sZ\!=\!\dX \times \euclideanspace_{+}$ and the resulting time-invariant dynamical system is
\begin{equation}
\label{eq:timeinvariant_system}
    \dhe = 
    \frac{\partial \dhe}{\partial \tau} =
    u_\theta(\dhe) =
    \left[\begin{matrix}
        \dot{\bm{h}}_\tau\\
        \dot{s}_\tau
    \end{matrix}\right]
.
\end{equation}
LaSalle's invariance principle can then be used to construct a stable target vector field. \citet{sprague2024stable} defined the function
\begin{equation}
    H(\bm{z}_\tau,\bm{z}_1) = \frac{1}{2}(\bm{z}_\tau-\bm{z}_1)^\trsp \bm{A} (\bm{z}_\tau - \bm{z}_1),
\end{equation}
with $\bm{A}= \left( \begin{smallmatrix}
         \lambda_{\bm{h}} \bm{I} & \bm{0} \\
         \bm{0} & \lambda_{s}
        \end{smallmatrix} \right)$ and $\lambda_{\bm{h}}, \lambda_{s} \in \mathbb{R}_{+}$.
Under the conditions in~\eqref{eq:invariance_dxae}, the steepest-descent solution yields the target field
\begin{equation}
\begin{aligned}
\label{eq:targetvf_safm}
    u(\bm{z}_\tau, \bm{z}_1) &= -\nabla_{\bm{z}} H(\bm{z}_\tau,\bm{z}_1)^\trsp \\ &= -\bm{A}(\bm{z}_\tau-\bm{z}_1) = 
    \begin{bmatrix}
    \lambda_{\bm{h}}(\bm{h}_1 - \bm{h}_{\tau}) \\
    \lambda_{s}(s_{1} - s_\tau)
    \end{bmatrix},
    \end{aligned}
\end{equation}
with $s_{0}=0$ and $s_{1}=1$. 
Note that, unlike~\eqref{eq:gausstargetvf}, the target vector field~\eqref{eq:targetvf_safm} does not have constant speed; instead, its magnitude decreases as the system approaches the target state $\bm{z}_1$. 
By using $s_\tau$ as the path parameter, the augmented state path induced by~\eqref{eq:targetvf_safm} can be written as
\begin{equation}
\label{eq:stable_gauss_path}
\bm{z}_{\tau}(s_\tau)
=
\begin{bmatrix}
\bm{h}_{\tau}(s_\tau)
\\
s_\tau
\end{bmatrix}
=
\begin{bmatrix}
\bm{h}_1 +
\left(
1 - s_\tau
\right)^{\frac{\lambda_{\bm{h}}}{\lambda_s}}
(\bm{h}_0 - \bm{h}_1)
\\[0.8em]
s_\tau
\end{bmatrix},
\end{equation}
which is used to sample points along the probability path employed during flow-matching training.
Convergence towards $\bm{z}_1$ by $\tau\!=\!1$ can then be enforced by choosing $\lambda_{\bm{h}}$ and $\lambda_s$ such that
\vspace{-0.2cm}
\begin{equation}
\label{eq:lambda_property}
    \lambda_s \geq -\ln\left(\epsilon_s\right),
    \quad
    \lambda_{\bm{h}} \geq -\ln\left(\frac{\epsilon_{\bm{h}}}{\|\bm{h}_0 - \bm{h}_1\|}\right),
\end{equation}
where $\epsilon_s, \epsilon_{\bm{h}}\in \euclideanspace_{+}$ are small constants with $d(1, s_{1}) < \epsilon_{s}$ and $d(\bm{h}_{1}, \hat{\bm{h}}_{1}) < \epsilon_{\bm{h}}$, and $\hat{\bm{h}}_{1}$ is the result of integrating $u_{\theta}$ over $[0,1]$. Given that at inference, the quantity $\|\bm{h}_0 - \bm{h}_1\|$ is unknown,~\citet{sprague2024stable} propose to first define $\lambda_{s}$ given some predefined $\epsilon_s$ and then set $\lambda_{\bm{h}} = \alpha\lambda_{s}$, where $\alpha \in \euclideanspace_{+}$ controls the convergence speed of $\bm{h}_{\tau}$ relative to $s_{\tau}$.
The conditional flow matching loss is finally constructed as 
\begin{equation}
\label{eq:CFM_loss_stable}
    \ell_{\text{CFMS}}= \mathbb{E}_{\substack{s_{\tau} \sim \text{Unif}(0, 1),\\ \bm{h}_0 \sim p_0,\, \bm{h}_1 \sim p_1}}
    \bigl\| u_\theta(\bm{z}_\tau) - \bm{A}(\bm{z}_1-\bm{z}_\tau) \bigr\|^2,
\end{equation}
with $\bm{h}_0 \sim \sN(\bm{0}, \bm{I})$.

\subsection{Lie Groups}
In robotics, learned dynamical systems are typically required to control both positions and orientations. The latter are represented as elements of the space of quaternions, corresponding to the hypersphere $\sA^3$, or of the special orthogonal group $\mathrm{SO}(3)$. 
Both are instances of a Lie group, which we introduce next. 

A Lie group $\mathcal{G}$ is a smooth manifold with a group structure. Specifically, it is endowed with a composition operation that is closed, associative, has an identity element, and holds an inverse to each element~\citep{Lee13:SmoothManifolds}. The smoothness of the manifold implies the existence of a tangent space $\sT_{\bm{x}}\sX$ at each state $\bm{x}_t \in \sX$, consisting of all the derivatives of curves on the manifold at $\bm{x}_t$. A dynamical system on a Lie group $\mathcal{X}$ is described as a function that assigns to each $\bm{x}_{t} \in \sX$ a velocity vector $\dot{\bm{x}}_{t} \in \sT_{\bm{x}}\sX$. The dynamics are therefore described in the disjoint union of all tangent spaces to the manifold, known as the tangent bundle $\sT\sX$.
A key property of Lie groups is their so-called \emph{Lie algebra} $\mathfrak{g}$, which is the tangent space at the group identity. Elements of the Lie algebra are converted onto elements of the group via the exponential map $\text{exp}:\mathfrak{g}\to\mathcal{G}$. The inverse operation is the logarithmic map $\text{log}:\mathcal{G}\to\mathfrak{g}$. Moreover, velocities can equivalently be represented as elements of the Lie algebra, $\dot{\bm{x}}_{t} \in \mathfrak{g}$, where a tangent vector expressed in the frame at $\bm{x}_t$ can be mapped to its representation in the frame at the identity via the \emph{adjoint} operator ${\textrm{Ad}_{\bm{x}}: \mathfrak{g} \to \mathfrak{g}}$, a linear map commonly written as a matrix. In the case of $\mathrm{SE}(3)$, the adjoint relates the body- and spatial-frame representations of velocities.

%% file: Sections/03_Problem_Formulation.tex
\section{Stable Flow Matching Dynamical Systems}
\label{sec:method_a}
We introduce \emph{Stable Flow Matching Dynamical Systems} (SFMDS), a framework that combines the expressivity of flow matching with formal stability guarantees for robot motion generation. This section introduces the core formulation, where the dynamics are implicitly modeled through a flow matching process conditioned on system states. Secs.~\ref{sec:enforcing_inductive_biases} and~\ref{sec:method-stability-enforcement} then describe how stability is enforced by constraining the learned flow to admissible velocity sets derived from the stability properties introduced in Sec.~\ref{subsec:dynamics-object}.

\subsection{Flow Matching Dynamical Systems}

We aim to learn probabilistic dynamical systems defined via a vector field $f$ of the form
\begin{equation}
\label{eq:conditioned_dynsym}
    \dot{\bm{x}}_t = f(\bm{x}_t, \bm{\omega}_t), \quad \text{with} \quad \bm{\omega}_t \sim \mathcal{N}(\bm{0}, \bm{I}),
\end{equation}
where the vector field is conditioned on a stochastic input $\bm{\omega}_t$.
We assume a given set of $N$ demonstrated trajectories $\kappa$ corresponding to rollouts containing position-velocity pairs of a target dynamical system as in~\eqref{eq:conditioned_dynsym}.
Our objective is to use these demonstrations to learn a parametric function $f_{\theta}$ that approximates $f$, where we require $f_{\theta}$ to belong to a subset of admissible functions $\admfuncset_\text{A}$ exhibiting a predefined stable behavior. By viewing each demonstrated state $\bm{x}_t$ as a sample from the state distribution $p_f(\bm{x}_t)$ induced by $f$, and each corresponding velocity as a sample from the target conditional velocity distribution $p_{f}(\dot{\bm{x}}_t  \!\mid \! \bm{x}_t)$, we aim to optimize 
\begin{equation}
\label{eqn:problem_formulation}
\begin{aligned}
\argmin_{\theta} \; 
&\mathbb{E}_{\bm{x}_t \sim p_f(\bm{x}_t)}
\ D_{\mathrm{KL}}\big(
p_{f}(\dot{\bm{x}}_t \! \mid \! \bm{x}_t)
\parallel
p_{\theta}(\dot{\bm{x}}_t \! \mid \! \bm{x}_t)
\big) \\
&\text{s.t.} \quad 
f_\theta(\bm{x}_t, \bm{\omega}_t) \in \admfuncset_{\text{A}} ,
\end{aligned}
\end{equation}
where $p_{\theta}(\dot{\bm{x}}_t \! \mid \! \bm{x}_t)$ denotes the velocity distribution induced by $f_\theta$ and $D_{\mathrm{KL}}$ is the (forward) Kullback-Leibler divergence.

We propose to model $f_{\theta}$ as the output of a flow trained via flow matching. Concretely, for each state $\bm{x}_{t}$, we compute a velocity $\dot{\bm{x}}_t$ using the flow map ${\phi_{\theta}:\dot{\mathcal{X}}\times[0,1]\times\mathcal{X} \to \dot{\mathcal{X}}}$ obtained by integrating the state-conditioned \emph{time-invariant} flow matching vector field $u_\theta(\bm{z}_\tau;\bm{x}_t)$ (Sec.~\ref{sec:stable_vector_fields}) over the auxiliary time $\tau \in [0, 1]$ as 
\begin{equation}
\begin{aligned}
\dot{\bm{x}}_t = f_{\theta}(\bm{x}_{t}, \bm{\omega}_t)
&= \phi_{\theta}(\bm{h}_0=\bm{\omega}_t, s_0=0, \tau=1;\bm{x}_{t}) = \bm{h}_1 ,\\
&= \underbrace{\begin{bmatrix} \bm{\omega}_t \\ 0 \end{bmatrix}}_{\bm{z}_0} + \int_{0}^{1}u_{\theta}(\bm{z}_\tau;\bm{x}_{t})\,d\tau.
\end{aligned}
\label{eq:stochastic_dynsym}
\end{equation}
The flow's initial condition is ${\bm{z}_0=[\bm{h}_0, s_0]^\trsp=[\bm{\omega}_{t}, 0]^\trsp}$.

Following the flow matching framework, we learn $u_{\theta}$ such that $f_\theta$ is a solution to~\eqref{eqn:problem_formulation} as described in the next section. The main challenge then lies in enforcing ${f_\theta \in \admfuncset_\text{A}}$, which requires \emph{(1)} identifying the admissible functions $\admfuncset_\text{A}$ (Sec.~\ref{sec:method-stability-enforcement}), and \emph{(2)} constraining $u_\theta$ so that $f_\theta\in\admfuncset_\text{A}$ (Sec.~\ref{sec:enforcing_inductive_biases}).
While our framework is general, we focus on enforcing two properties: \emph{(1)} positive invariance, and \emph{(2)} asymptotic stability, leading to the proposed \ac{sfmds}. 

%% file: Sections/04_Method_imitation_learning.tex
\subsection{Imitation Learning}
\label{app:imitation-learning}

We frame the problem in~\eqref{eqn:problem_formulation} in the context of imitation learning, where the goal is to learn the dynamics $f$ from demonstrated trajectories. Here, we describe how to train flow matching dynamical systems using a behavior cloning objective, before enforcing ${f_\theta \in \admfuncset_\text{A}}$ in the next section. 

\subsubsection{Predefined Pseudo-Time Dynamics}
\citet{sprague2024stable} learn the dynamics of the full time-invariant system in~\eqref{eq:timeinvariant_system}, i.e., both $\dot{\bm{h}}_\tau$ and $\dot{s}_\tau$. However, the dynamics of the pseudo-time $s_\tau$ are fully specified beforehand as the target state $s_{1} = 1$ is fixed and its desired behavior is known a priori. Therefore, we split $u_\theta$ into two components $u_{\theta}^{h}$ and $u^{s}$, as
\begin{equation}
    \dhe = 
    u_\theta(\he; \bm{x}_{t}) =
        \left[\begin{matrix}
        u^{h}_\theta(\bm{h}_\tau, s_\tau;\bm{x}_t)\\
        u^{s}(s_\tau)
    \end{matrix}\right] =
    \left[\begin{matrix}
        \dot{\bm{h}}_\tau\\
        \dot{s}_\tau
    \end{matrix}\right]
,
\end{equation}
where $u^{h}_{\theta}$ is the only learnable component, while $u^{s}$ is fixed to the target dynamics in~\eqref{eq:targetvf_safm}, i.e.,
\begin{equation}
\label{eq:us}
u^{s} = \lambda_{s}(s_1 - s_\tau), \quad \text{with} \quad s_1 = 1, \ 
\lambda_s = -\ln\left(\epsilon_{s}\right).
\end{equation}

\begin{remark}
    From here on, we drop $h$ in $u^{h}_{\theta}$, writing simply $u_{\theta}$, as the remainder of the method concerns only this component.
\end{remark}

\subsubsection{Loss Function}
We minimize the velocity distribution divergence introduced in~\eqref{eqn:problem_formulation} through the flow matching loss~\eqref{eq:CFM_loss_stable}~\cite{su2025flow}, yielding
\begin{equation}
\label{eq:il}
    \ell_{\text{IL}}=\mathbb{E}_{\!\! \substack{s_{\tau}\sim \text{Unif}(0,1),\,\bm{h}_0 \sim p_0\\\bm{x}_t\sim p_{f}(\bm{x}_t), \dot{\bm{x}}_{t}\sim p_f(\dot{\bm{x}}_t \mid \bm{x}_{t})}} \lVert u_{\theta}\left(\bm{h}_\tau, s_{\tau};\bm{x}_t\right) - \lambda_{\bm{h}}(\dot{\bm{x}}_{t} - \bm{h}_{\tau})\rVert^{2}.
\end{equation}
Similarly to previous works~\citep{perez2023stable,xie2023neural}, we mitigate the distribution shift inherent to such \emph{one-step} imitation learning objectives by optimizing the policy parameters $\theta$ to minimize a \emph{multi-step} prediction error over rollouts of the learned dynamics. In this work, we use the forward Euler integration update $\hat{\bm{x}}_{t+1}=\hat{\bm{x}}_t + f_{\theta}(\hat{\bm{x}}_{t}, \bm{\omega}_t)\Delta t$, where $\hat{\bm{x}}_{t}$ denotes states visited by following $f_{\theta}$, $\hat{\bm{x}}_0 = \bm{x}_0$ is initialized from states belonging to the demonstrated trajectories,
and $\Delta t$ is the integration step size. To maintain the simulation-free training procedure of flow matching, we approximate the terminal flow state $\bm{h}_1$, i.e., the dynamics in~\eqref{eq:stochastic_dynsym}, using the approximation introduced in~\citet{ding2025fast}
\begin{equation}
    f_\theta(\hat{\bm{x}}_{t},\bm{\omega}_t)\approx \bm{h}_0 + u_\theta(\bm{h}_0, s_\tau; \hat{\bm{x}}_t) / \lambda_{\bm{h}}.
\end{equation}
This leads to the loss over predicted positions
\begin{equation}
\label{eq:il_bptt}  \ell^{\kappa}_{\text{IL}}=\mathbb{E}_{\substack{s_{\tau}\sim \text{Unif}(0,1),\,\bm{h}_0 \sim p_0 \\ (\bm{x}_0,\bm{x}_{t+1})\sim p_{f}(\bm{x}_0,\bm{x}_{t+1}), \\ \,\hat{\bm{x}}_{t}\sim p_{\theta}(\hat{\bm{x}}_{t}\mid\bm{x}_0)}}  \lVert \bm{x}_{t+1} - \hat{\bm{x}}_{t+1}(\bm{h}_0, s_{\tau}; \hat{\bm{x}}_t)\rVert^{2},
\end{equation}
computed for each integration step, where $p_f(\bm{x}_0,\bm{x}_{t+1})$ denotes the joint distribution over demonstrated initial states and their corresponding future states, and $p_{\theta}(\hat{\bm{x}}_t \! \mid \! \bm{x}_0)$ denotes the distribution over learned rollout states. The total loss is then defined by the sum of~\eqref{eq:il_bptt} over all integration steps.
Next, we discuss how to constrain flow matching dynamical systems to admissible sets, a prerequisite for enforcing stability properties on the learned dynamics.

%% file: Sections/05_Method_A.tex
\section{Constraining Flow Matching Dynamical Systems to Admissible Sets}
\label{sec:enforcing_inductive_biases}
In this section, we assume a given set of admissible functions $\mathcal{F}_{\text{A}}$ and we aim to learn a vector field ${f_\theta(\bm{x}_t, \bm{\omega}_t) \in \mathcal{F}_{\text{A}}}$
by constraining its output $\dot{\bm{x}}_t$ to a corresponding set of admissible velocities $\dXaonly \subseteq \dX$, see Fig.~\ref{fig:summary_admissible_sets}(a). To achieve this, we require the state $\bm{h}_\tau$ of the flow matching vector field to converge to $\dXaonly$ by $\tau = 1$, thus ensuring that $\dot{\bm{x}}_t = \bm{h}_1 \in \dXaonly$, as shown in Fig.~\ref{fig:summary_admissible_sets}(b). This can be expressed as satisfying
\begin{equation}
\label{eq:convergence_dXa}
d\big(\bm{h}_1, \dXaonly\big) = 0, \quad \forall \bm{h}_{0} \in \dX, \ \forall \bm{x}_t \in \mathcal{X},
\end{equation}
with $d\big(\bm{h}_\tau, \dXaonly\big) \!=\! \inf_{\dot{\bm{x}}_{\textnormal{a}} \in \dXan} d(\bm{h}_\tau, \dot{\bm{x}}_{\textnormal{a}})$. 
Crucially, since the evolution of $\bm{h}_\tau$ is governed by the learned $u_\theta$, enforcing the desired behavior on $\bm{h}_{\tau}$ translates into constraining $u_\theta$ accordingly. Hence, we define a subset of admissible values $\sU_{\text{A}} \subseteq \sU$, see Fig.~\ref{fig:summary_admissible_sets}(c). We describe how to constrain $u_\theta$ next.
\begin{remark}
    In general, the admissible sets $\dXaonly$ and $\dUaonly$ can depend on the current states $\bm{x}_t$ and $\bm{h}_{\tau}$, as well as on the learning parameters $\theta$, i.e., $\dXa$ and $\dUa$.
\end{remark}
\subsection{Achieving admissible velocities}
\label{sec:enforice_stability_u}
Taking inspiration from~\citet{sprague2024stable}, we leverage LaSalle's invariance principle (see Sec.~\ref{sec:lasalle}) for imposing the convergence of $\bm{h}_\tau$ to the set $\dXaonly$.
We define the function $H(\he)$ as the squared distance to the admissible set $\dXaonly$,
\begin{equation}
\label{eq:H-func}
\begin{aligned}
H(\he) = d(\bm{h}_\tau, \dXaonly)^{2}
=\left\lVert \bm{\delta}_{\bm{h}} \right\rVert^2
\end{aligned}
\end{equation}
with
\begin{equation}
\label{eq:projection-delta-h}
\begin{aligned}
\bm{\delta}_{\bm{h}}
&= \Pi_{\dXaonly}(\bm{h}_{\tau}) - \bm{h}_\tau, \\
\Pi_{\dXaonly}(\bm{h}_{\tau})
&= \argmin_{\dot{\bm{x}}_{\textrm{a}} \in \dXaonly}
\left\lVert \bm{h}_{\tau} - \dot{\bm{x}}_{\textrm{a}} \right\rVert,
\end{aligned}
\end{equation}
where $\bm{\delta}_{\bm{h}}$ is the vector that projects $\bm{h}_\tau$ onto $\dXaonly$, and $\Pi_{\dXaonly}$ is the projection mapping~\cite[Ch.~8.1]{boyd2004convex}.
Substituting~\eqref{eq:H-func} in LaSalle's conditions~\eqref{eq:invariance_dxae} yields
\begin{equation}
\label{eq:condition-u}
\begin{aligned}
(\nabla_{\bm{h}} \left\lVert \bm{\delta}_{\bm{h}} \right\rVert^2)^\trsp u_{\theta} &< 0, \quad \forall \he \notin \dXaonly, \\
(\nabla_{\bm{h}} \left\lVert \bm{\delta}_{\bm{h}} \right\rVert^2)^\trsp u_{\theta} &= 0, \quad \forall \he \in \dXaonly,
\end{aligned}
\end{equation}
thus constraining $u_{\theta}$ such that $\bm{h}_{\tau} \in \dXaonly$ as $\tau \to \infty$. 
However, for flow matching dynamical systems, convergence to $\dXaonly$ must occur sufficiently fast, i.e., by $\tau=1$. We return to this finite-time requirement after deriving the close-form projection for the specific case where $\dXaonly$ is a half-space.

\begin{figure}[t]
    \centering
    \begin{overpic}[width=\linewidth]{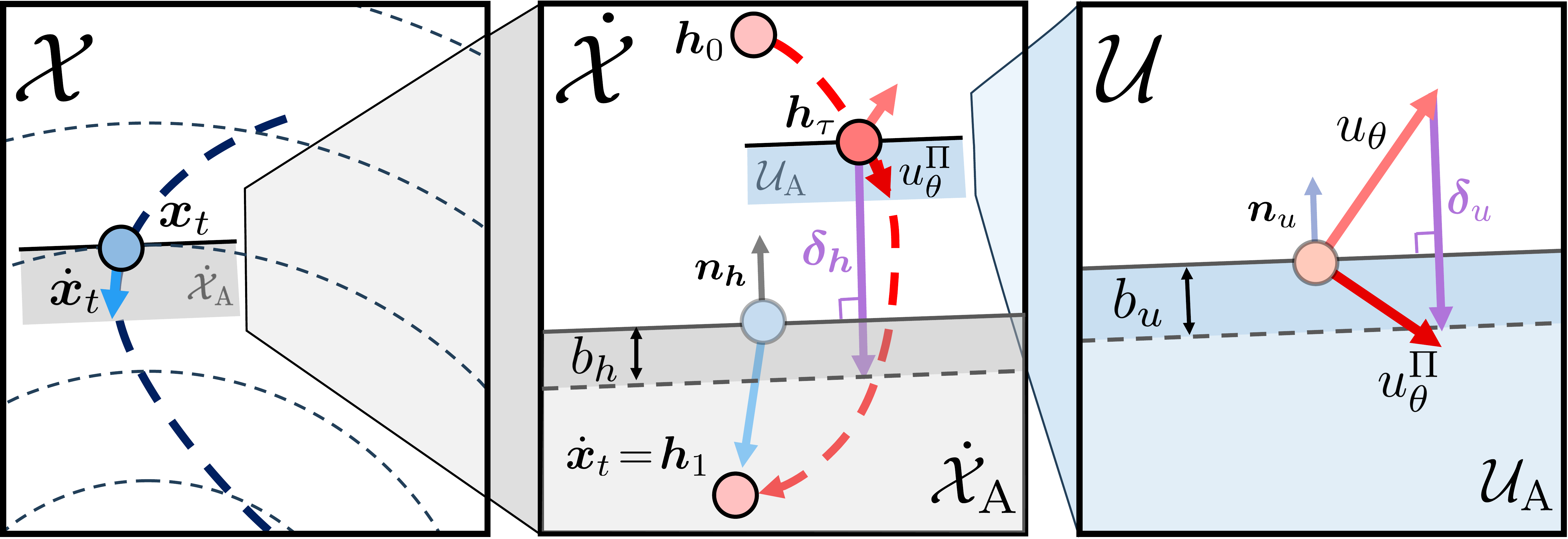}
        \put(13,-3){\footnotesize (a)}
        \put(47,-3){\footnotesize (b)}
        \put(81,-3){\footnotesize (c)}
    \end{overpic}
    \caption{Overview of SFMDS. (a) Velocities $\dot{\bm{x}}_t$ must belong to the half-space $\dXa$ (\grayrectangle). (b) The evolution of $\bm{h}_\tau$, governed by $u_\theta$, is constrained to achieve this condition by $\tau=1$, as $\bm{h}_1 = \dot{\bm{x}}_t$. (c) Hence, $u_\theta$ is constrained to belong to the admissible half-space $\sU_{\text{A}}$ (\graybluerectangle).}
    \label{fig:summary_admissible_sets}
\end{figure}

\textbf{Half-space case.} As will be discussed in Sec.~\ref{sec:method-stability-enforcement}, the admissible velocity sets $\dXaonly$ considered in this work, for continuous-time systems $f_{\theta}$, are half-spaces whose boundaries contain the origin. The projection to this set, depicted in Fig.~\ref{fig:summary_admissible_sets}(b), is given by~\cite[Ch.~8.1.1]{boyd2004convex}
\begin{equation}
\label{eq:proj-vector}
\bm{\delta}_{\bm{h}} = -\sigma_r(\bm{n}_{\bm{h}}^{\trsp}\bm{h}_{\tau}+m_{\bm{h}})\bm{n}_{\bm{h}}/\|\bm{n}_{\bm{h}}\|^{2},
\end{equation}
with the outward-pointing normal $\bm{n}_{\bm{h}} \in \euclideanspace^{n}$, a margin $m_{\bm{h}} \in \euclideanspace_{+}$, and the ReLU function $\sigma_r$.
The margin $m_{\bm{h}}\!=\!b_{\bm{h}}\|\bm{n}_{\bm{h}}\|$ is employed to ensure that the projected vector lies strictly inside the half-space, where $b_{\bm{h}}$ defines the shift of the half-space along $-\bm{n}_{\bm{h}}$, as depicted in Fig.~\ref{fig:summary_admissible_sets}(b). The ReLU function ensures that no projection is computed if $\bm{h}_{\tau}$ is already in $\dXaonly$. From \eqref{eq:proj-vector}, it follows that
\begin{equation}
\label{eq:gradient_half-space}
    \nabla_{\bm{h}} \left\lVert \bm{\delta}_{\bm{h}} \right\rVert^2 = -2\bm{\delta}_{\bm{h}}.
\end{equation}

\textbf{Convergence by $\tau=1$.} Next, we ensure convergence to $\dXaonly$ by $\tau=1$. To this end, we note that, similarly to $\dXaonly$, the conditions in~\eqref{eq:condition-u} define a half-space of admissible values $\dUaonly$ for $u_\theta$, whose boundary passes through the origin and whose normal direction is given by the vector~\eqref{eq:gradient_half-space}, as depicted in Fig.~\ref{fig:summary_admissible_sets}(c). Intuitively, this constrains the vector field $u_{\theta}$ so that the flow $\bm{h}_{\tau}$ approaches the set $\dXaonly$ as $\tau$ progresses. However, this condition still admits arbitrarily low rates of change for $\bm{h}_\tau$. We impose a minimum convergence rate by introducing the offset $b_{u}$ (see Fig.~\ref{fig:summary_admissible_sets}(c)) to the half-space $\dUaonly$ through a margin $m_u\!=\!b_{u} \| \bm{n}_u\| $, analogous to the margin in~\eqref{eq:proj-vector}. By introducing this margin and substituting~\eqref{eq:gradient_half-space} into~\eqref{eq:condition-u}, we obtain the convergence conditions to the half-space $\dXaonly$ 
\begin{equation}
\begin{aligned}
    \label{eq:def-mB}
    &-\bm{\delta}_{\bm{h}}^\trsp u_{\theta} + m_{u}< 0, \quad \forall \bm{h}_{\tau}\notin \dXaonly, \quad \\ 
    &-\bm{\delta}_{\bm{h}}^\trsp u_{\theta} = 0, \quad \forall \bm{h}_{\tau} \in \dXaonly .
\end{aligned}
\end{equation}
Since the normal vector $\bm{n}_{u}$ can be scaled by any positive constant without changing the half-space, we use the simpler choice $\bm{n}_u=-\bm{\delta}_{\bm{h}}$ instead of $\bm{n}_u=-2\bm{\delta}_{\bm{h}}$.

We proceed to determine the value of $m_u$ required to ensure convergence to $\dXaonly$ by $\tau\!=\!1$, following the convergence criterion introduced in Sec.~\ref{sec:stable_vector_fields}, see~\eqref{eq:targetvf_safm}-\eqref{eq:lambda_property}. Specifically, we note that the offset $b_u$ lower bounds the component of $u_\theta$ along the steepest-descent direction of the function $H$, which determines the rate at which $\bm{h}_\tau$ approaches $\dXaonly$. Since the stable autonomous target field $u = \lambda_{\bm{h}}(\bm{h}_1 - \bm{h}_\tau)$ in~\eqref{eq:targetvf_safm} prescribes motion along this same direction, we use its convergence criterion to define $b_u$. In particular, the magnitude of $u$ is proportional to the distance to the target. Here, the relevant target point is the closest point on the boundary of the admissible half-space obtained by projecting $\bm{h}_\tau$ onto $\dXaonly$, which is at distance $\|\bm{\delta}_{\bm{h}}\|$ from $\bm{h}_\tau$. Therefore, convergence to $\dXaonly$ is achieved by $\tau\!=\!1$ if
\begin{equation}
    b_u \geq \lambda_{\bm{h}}\|\bm{\delta}_{\bm{h}}\|.
\end{equation}
Choosing the minimal admissible offset, we obtain the margin
\begin{equation}
    m_u = b_u\|\bm{n}_u\| =  b_u\|\bm{\delta}_{\bm{h}}\|
    = \lambda_{\bm{h}} \|\bm{\delta}_{\bm{h}}\|^2 .
\end{equation}

\subsection{Constraining $u_\theta$ to the admissible space $\mathcal{U}_{\textrm{A}}$}
\label{sec:constraining_u}
We now have all the necessary elements for constraining $u_{\theta}$ to the half-space $\dXaonly$. As any $u_\theta$ fulfilling the half-space conditions~\eqref{eq:def-mB} guarantees convergence, we allow the flow matching vector field to select any such $u_{\theta}$ while minimizing the imitation learning loss~\eqref{eq:il_bptt}. We note that the second condition in~\eqref{eq:def-mB} is readily fulfilled by construction via the ReLU function $\sigma_r$ in~\eqref{eq:proj-vector}. Hence, it remains to enforce the first condition, which we achieve through two variants: a \emph{soft constraint} via a penalty term, and a \emph{hard structural constraint} imposed into the model. 

\subsubsection{Soft-constrained $u_{\theta}$}
\label{sec:data-driven-u}
Positive quantities for the left-hand side of~\eqref{eq:def-mB} can be directly penalized via the \emph{hinge-type loss}
\begin{equation}
\label{eq:stability_loss}
\ell_{\text{PI}}=\max(0, -\bm{\delta}_{\bm{h}}^\trsp u_{\theta} + m_{u})=\sigma_{r}(-\bm{\delta}_{\bm{h}}^\trsp u_{\theta} + m_{u}).
\end{equation}
The loss~\eqref{eq:stability_loss} must be minimized $\forall \bm{h}_{\tau} \in \dot{\mathcal{X}}$ and $\forall \bm{x}_t \in\mathcal{X}$. In practice, we sample uniformly from $\dot{\mathcal{X}}$ and $\mathcal{X}$ during training.

\subsubsection{Hard-constrained $u_{\theta}$}
Alternatively, we strictly constrain $u_{\theta}$ to satisfy~\eqref{eq:def-mB}, by projecting $u_{\theta}$ onto $\dUaonly$, analogously to~\eqref{eq:proj-vector}. We obtain the projected vector field
\begin{equation}
\label{eq:proj-vect-u-stable}
\begin{aligned}
&u^{\Pi}_{\theta}=\Pi_{\dUan}(u_{\theta}) = u_{\theta} + \bm{\delta}_u, \\
\text{with} \quad &\bm{\delta}_u=-\sigma_r\left(\bm{n}_u^{\trsp}u_{\theta} + m_{u}\right)\bm{n}_u / \|\bm{n}_u\|^{2},
\end{aligned}
\end{equation}
see Fig.~\ref{fig:summary_admissible_sets}(c).
Hard constraints are architecturally more restrictive than soft constraints, in exchange for provable guarantees.

%% file: Sections/06_Method_B.tex
\section{Enforcing Stability}
\label{sec:method-stability-enforcement}
In the previous section, we introduced constrained flow matching dynamical systems, whose learned vector field $u_\theta$ generates velocities in a given admissible set $\dXaonly$. Next, we identify admissible sets for two stability properties.

\subsection{Positively-invariant Flow Matching Dynamical Systems}
\label{sec:positive_invariance}
Positive invariance is crucial when the state of a dynamical system $f_{\theta}$ must remain within a specified region $\sA$, e.g., a robot’s workspace. It is also relevant when minimizing penalty losses, e.g.,~\eqref{eq:stability_loss}, over a state space, as the problem must be constrained to a closed region to be tractable.

Similarly to~\cite{lemme2014neural, perez2024puma}, we enforce positive invariance by defining the set of admissible velocities as 
\begin{equation}
\label{eq:positive_invariant_set}
    \dXaonly(\bm{x}_t)  = \left\{\dot{\bm{x}}_{t} \in \dot{\sX} \mid  \bm{n}_{\partial}^\trsp  \dot{\bm{x}}_{t} \leq 0\right \}, \quad \forall \bm{x}_{t} \in \partial\sA,
\end{equation}
where $\bm{n}_{\partial}$ is the outward-pointing normal vector at the boundary $\partial\sA$ of $\sA$.
We introduce \emph{positively-invariant flow matching dynamical systems} as flows constrained to the set~\eqref{eq:positive_invariant_set} at every state $\bm{x}_{t} \in \partial\sA$.
This is achieved by constraining the vector field $u_{\theta}$ as discussed in Sec.~\ref{sec:enforcing_inductive_biases}.
In this case, we use the normal vector $\bm{n}_{\bm{h}}\!=\!\bm{n}_{\partial}$ and set $m_{\bm{h}}=0$ to compute the projection $\bm{\delta}_{\bm{h}}$~\eqref{eq:proj-vector}, imposing then the desired constraints either as a soft penalty~\eqref{eq:stability_loss} or as strict constraint~\eqref{eq:proj-vect-u-stable}.

\subsection{Asymptotically-stable Flow Matching Dynamical Systems}
\label{subsec:asymptstable_FMDS}
For goal-driven tasks, we define the admissible velocity set $\dXaonly$ using Lyapunov conditions, so that the target pose $\xe \!\in\! \sX$ becomes an asymptotically stable equilibrium of $f_{\theta}$.
While Lyapunov stability theory typically requires $f_{\theta}$ to be time-differentiable, flow matching dynamical systems~\eqref{eq:stochastic_dynsym} define a distribution over $\dot{\bm{x}}_{t}$, rendering them non-differentiable, similarly to stochastic differential equations~\citep{oksendal2013stochastic}. Instead, we perform our stability analysis on  \emph{sampled trajectories}, ensuring time-differentiability. 
These trajectories are obtained by sampling an initial $\bm{\omega}_0 \sim \mathcal{N}(\mathbf{0}, \mathbf{I})$ and subsequently evolving $\bm{\omega}$ deterministically. This can be achieved, e.g., by making $\bm{\omega}_t$ track $\dot{\bm{x}}_t$ through the first-order dynamics $\alpha\dot{\bm{\omega}}_t = \dot{\bm{x}}_t - \bm{\omega}_t$, with $\alpha > 0$. As $\alpha \to \infty$, $\bm{\omega}_t$ remains constant at $\bm{\omega}_0$; as $\alpha \to 0$, it converges instantaneously to $\dot{\bm{x}}_t$. Intuitively, by sampling only $\bm{\omega}_{0}$, we select one of the infinitely-many dynamical systems encoded by the flow matching model.

\subsubsection{Stability conditions}

Inspired by~\cite{Rana2020:EuclideanizingFlows, Urain20:ImitationFlow, perez2024puma}, we enforce stability w.r.t. a Lyapunov function in a latent space $\mathcal{L}\subseteq \mathbb{R}^{m}$ defined via a neural network $\psi_{\theta}: \mathcal{X} \to \mathcal{L}$, where a simple latent Lyapunov function can be identified with an arbitrary complex Lyapunov function in the original space $\mathcal{X}$. We define latent states as $\bm{y}_t \!=\! \psi_{\theta}(\bm{x}_t)$, yielding the latent equilibrium $\bm{y}_e \!=\! \psi_{\theta}(\bm{x}_e)$ and latent dynamics
\begin{equation}
\label{eq:latent_dynamics}
\dot{\bm{y}}_{t}=\frac{\partial \psi_{\theta}}{\partial t} (\bm{x}_t)=\bm{J}_{\psi_{\theta}}(\bm{x}_{t})f_{\theta}(\bm{x}_{t},\bm{\omega}_{t}),
\end{equation}
where $\bm{J}_{\psi_{\theta}}$ denotes the Jacobian of $\psi_{\theta}$ with respect to $\bm{x}_{t}$. To enforce asymptotic stability, two challenges must be addressed, which we achieve through two stability conditions. First, if we construct a valid Lyapunov candidate~\eqref{eq:lyapunov_1} in $\sL$, we must ensure that this candidate remains valid once expressed in $\sX$. Second, we must ensure that the value of this function decreases as time evolves, making this candidate a valid Lyapunov function~\eqref{eq:lyapunov_2} and thus ensuring asymptotic stability.
We first construct a simple Lyapunov candidate using the Euclidean distance
\begin{equation}
    V^{\mathcal{L}}(\bm{y}_t) = d(\ye, \bm{y}_{t})^{2}
\end{equation}
in $\mathcal{L}$ leading to 
\begin{equation}
    V^{\mathcal{X}}(\bm{x}_t) = d(\psi_{\theta}(\xe),\psi_{\theta}(\bm{x}_t) )^{2}
\end{equation}
in $\sX$. A valid candidate must be \emph{positive definite}~\eqref{eq:lyapunov_1}, which $V^{\mathcal{L}}$ is by construction. While $V^{\mathcal{X}}$ is non-negative, it can, for an arbitrary $\psi_{\theta}$, equal $0$ for some $\bm{x}_t \neq \xe$, thus breaking~\eqref{eq:lyapunov_1}. As a result, we require the following injectivity condition at $\xe$~\cite{perez2024puma}:
\begin{condition}
\label{cond:lyap_candidate}
$V^{\sX}$ is a valid Lyapunov candidate if ${\psi_{\theta}(\bm{x}_t) = \ye \iff \bm{x}_t = \xe}$, $\forall \bm{x}_t \in \sX$.
\end{condition}
Given a valid candidate, previous approaches~\citep{Rana2020:EuclideanizingFlows,Urain20:ImitationFlow,perez2023stable} enforce the decrease condition~\eqref{eq:lyapunov_2} by constraining $\dot{\bm{y}}_{t}$ to align with the steepest descent direction of $V^{\sL}$. This guarantees $\dot{V}^{\sL} < 0$ and, consequently, ${\dot{V}^{\sX} < 0}$.
However, constraining the network to align with one of the infinitely many admissible solutions can be too restrictive when learning multimodal behaviors.
Instead, we propose to constrain $\dot{\bm{y}}_t$ to lie in the admissible latent set $\dXaLonly$ satisfying $\dot{V}^{\sL} < 0$,
\begin{equation}
\label{eq:latent_ad_set}
    \dXaL = \{\dot{\bm{y}}_{t} \in \dXL \mid \nabla_{\bm{y}} V^{\mathcal{L}}(\bm{y}_t)^\trsp \dot{\bm{y}}_{t} < 0 \}.
\end{equation}
Given the decrease condition in~\eqref{eq:lyapunov_2}, constraining the model to such a set resembles the approach of~\citet{kolter2019learning}, although they do not consider latent Lyapunov functions.
Using~\eqref{eq:latent_dynamics}, we can construct the corresponding set in $\dot{\mathcal{X}}$ as 
\begin{equation}
    \label{eq:admissible_vel_set_stable}
    \dXa  = \{ \dot{\bm{x}}_{t} \in \dot{\mathcal{X}} \mid \nabla_{\bm{x}} V^{\mathcal{X}}(\bm{x}_t)^\trsp \bm{J}_{\psi_{\theta}}\, \dot{\bm{x}}_{t} < 0 \}.
\end{equation}
As computing $\dot{\mathcal{X}}_{\text{A}}$ requires evaluating $\nabla_{\bm{y}} d(\ye, \bm{y}_t)$ in $\sL$, we work directly with $\dot{\mathcal{L}}_{\text{A}}$, leading to the following asymptotic stability condition:
\input{Algorithms/soft_SFMDS}
%
\begin{condition}
\label{cond:lyap_function}
Given a valid Lyapunov candidate $V^{\sX}$, $\xe$ is an asymptotically stable equilibrium of $f_\theta$ in $\sX$ if $\dot{\bm{y}}_t \in \dXaLonly$,  $\forall \bm{x}_t \in \sX \setminus \{\xe\}$.
\end{condition}
\footnotetext{Loss described for single-step scenario.}
Note that this aligns with the conditions in~\cite[Thm 3]{perez2024puma}.
\subsubsection{Enforcing latent constraints}
We introduce \emph{Stable Flow Matching Dynamical Systems} (SFMDS) as constrained flow matching dynamical systems over the set~\eqref{eq:admissible_vel_set_stable}, or equivalently, the latent set~\eqref{eq:latent_ad_set}. \ac{sfmds} constrains a latent vector field $\uL$ to generate velocities $\dot{\bm{y}}_{t}\in\dXaLonly$ as follows. 
We define the latent flow dynamics 
\begin{equation}
    \frac{\partial \hL}{\partial \tau} = \uL(\hL, s_{\tau}; \bm{x}_{t})
\end{equation}
with $\hL \!\in\! \dXL$, so that $\dot{\bm{y}}_{t}\!=\!\tilde{\bm{h}}_1$.
From~\eqref{eq:latent_dynamics}, we can relate flow states in $\sX$ and $\sL$ as
$\hL = \bm{J}_{\psi_{\theta}}\bm{h}_{\tau}$, and thus
\begin{equation}
\label{eq:u_relation}
    \uL =  \frac{\partial \hL}{\partial \tau} = \frac{\partial (\bm{J}_{\psi_{\theta}}\bm{h}_{\tau})}{\partial \tau} = \bm{J}_{\psi_{\theta}}\frac{\partial \bm{h}_{\tau}}{\partial \tau} = \bm{J}_{\psi_{\theta}} u_{\theta},
\end{equation}
where we used that $\bm{J}_{\psi_{\theta}}$ is constant w.r.t. $\tau$.
Finally, we enforce $\dot{\bm{y}}_{t}\!=\!\tilde{\bm{h}}_1\!\in\!\dXaLonly$ via soft or hard constraints, as described in Sec.~\ref{sec:enforcing_inductive_biases} and elaborated next.

\paragraph{Soft \ac{sfmds}}
\label{subsec:data_driven_stability}
Using~\eqref{eq:u_relation}, we reformulate the hinge-type loss~\eqref{eq:stability_loss} in the latent space as
\begin{equation}
\label{eq:stability_loss_sfmds}
\ell_{\text{AS}} = \sigma_{r}\big(-\bm{\delta}_{\tilde{\bm{h}}}^\trsp \uL + \tilde{m}_u \big) = \sigma_{r}\big(-\bm{\delta}_{\tilde{\bm{h}}}^\trsp \bm{J}_{\psi_{\theta}} u_{\theta} + \tilde{m}_u \big),
\end{equation}
where $\bm{\delta}_{\tilde{\bm{h}}}$ is the vector that projects $\tilde{\bm{h}}_{\tau}$ onto $\dXaLonly$, and $\tilde{m}_u$ is the latent margin. From \eqref{eq:latent_ad_set}, the outward-pointing normal vector of $\dXaLonly$ is $\nabla_{\bm{y}_{t}} d(\ye, \bm{y}_{t})^2$, from which we compute $\bm{\delta}_{\tilde{\bm{h}}}$ following~\eqref{eq:proj-vector}. Note that there is no restriction on the dimensionality of $\bm{y}_t$, enabling $\dim(\sL) \neq \dim(\sX)$ for $\psi_{\theta}$.

Minimizing~\eqref{eq:stability_loss_sfmds} enforces that the latent dynamics $\dot{\bm{y}}_t$~\eqref{eq:latent_dynamics} belong to $\dXaLonly$ (Cond.~\ref{cond:lyap_function}). 
Importantly, it also implicitly enforces $\psi_{\theta}$ to be injective at the equilibrium (Cond.~\ref{cond:lyap_candidate}): The latent margin $\tilde{m}_{\bm{h}}$ in~\eqref{eq:proj-vector} enforces $\dot{\bm{y}}_t \neq 0$ with ${\bm{y}_t \!=\! \psi_{\theta}(\bm{x}_t)}$, $\forall \bm{x}_t \!\in\! \sX \backslash \{\bm{x}_\text{e}\}$, implying $\bm{y}_t \neq \bm{y}_{\text{e}}$ as $\dot{\bm{y}}_\text{e} = 0$ at the equilibrium point, as similarly described in~\cite{perez2024puma}.

In practice, the loss~\eqref{eq:stability_loss_sfmds} can only be tractably enforced in a closed region, i.e., $f_{\theta}$ must always stays within this region~\citep{lemme2014neural}. This is achieved by integrating the loss~\eqref{eq:stability_loss} for positive invariant sets derived from~\eqref{eq:positive_invariant_set}. 
As a result, soft \ac{sfmds} imposes \emph{local} asymptotic stability. A summary of this method is provided in Algorithm~\ref{algo:soft-sfmds}.
\begin{figure}[t]
    \centering
    \includegraphics[width=\linewidth,trim=0 170 200 185,clip]{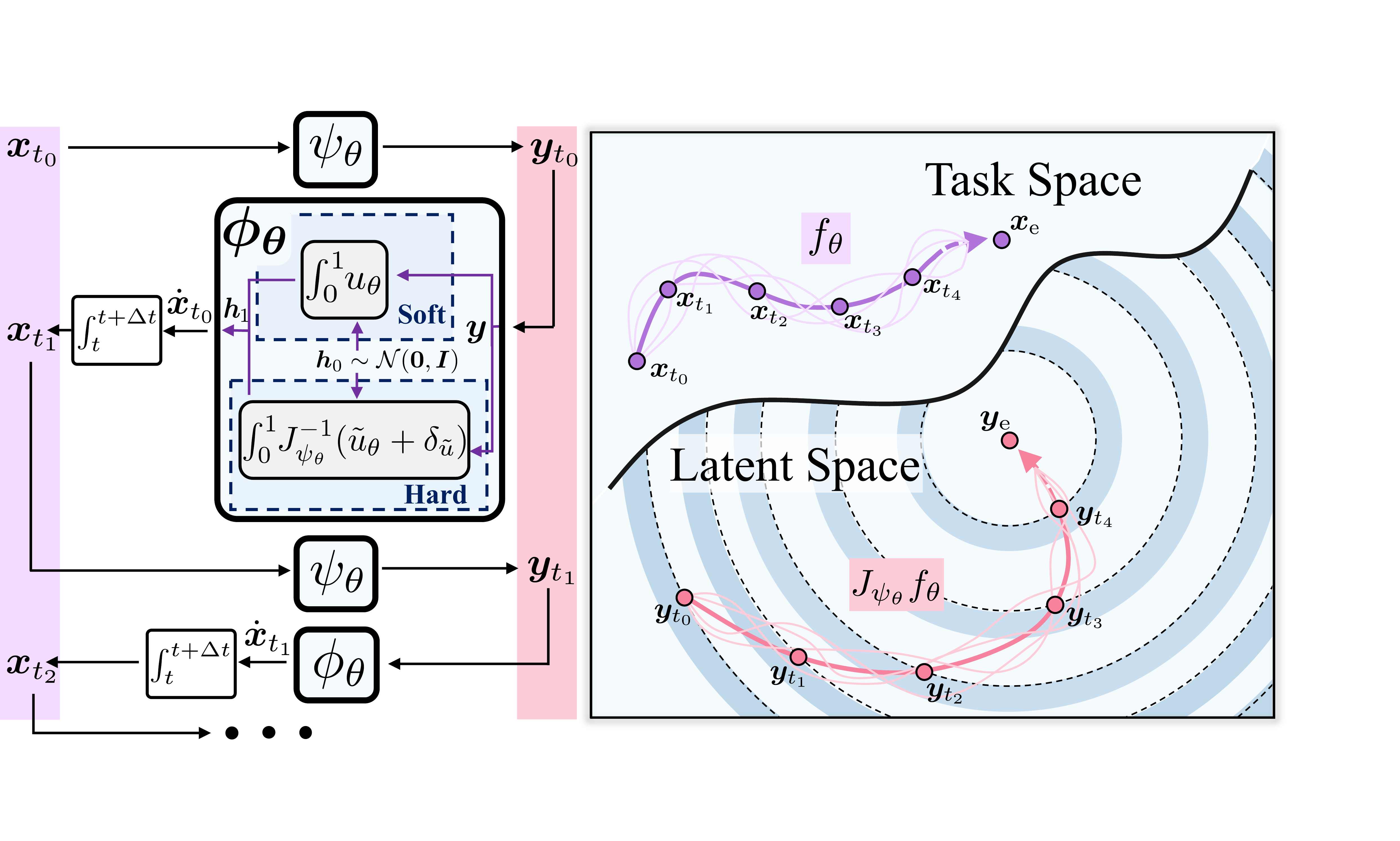}
    \caption{2D illustration of trajectory inference for soft/hard SFMDS. Each forward Euler step moves $\bm{y}_t$ closer to $\bm{y}_{\textrm{e}}\!=\!\psi_{\theta}(\bm{x}_\textrm{e})$, thereby $\bm{x}_t$ closer to $\bm{x}_{\textrm{e}}$.}
    \label{fig:method_2d_example}
\end{figure}
%
\input{Algorithms/hard_SFMDS}
\paragraph{Hard \ac{sfmds}}
\label{subsec:hard_stability}
In contrast to penalty losses, strict constraints can guarantee asymptotic stability \emph{globally} by construction. As described next, in this case, we require $\bm{J}_{\psi_{\theta}}$ to be invertible, implying that $\psi_{\theta}$ must be a diffeomorphism~\cite{Rana2020:EuclideanizingFlows}. The bijectivity of such a mapping fulfills Cond.~\ref{cond:lyap_candidate} by construction, while also enforcing $\dim(\sL) = \dim(\sX)$. Akin to~\eqref{eq:proj-vect-u-stable}, we project the latent vector field $\tilde{u}_{\theta}$ onto the half-space $\dXaLonly$ (Cond.~\ref{cond:lyap_function}) and obtain the latent projected vector field
\begin{equation}
    \label{eq:projection_hard_sfmds}
    \begin{aligned}
    \tilde{u}_{\theta}^{\Pi} = \tilde{u}_{\theta} + \bm{\delta}_{\tilde{u}}, 
    \text{ with } \bm{\delta}_{\tilde{u}} = \sigma_r\left(-\bm{\delta}_{\tilde{\bm{h}}}^{\trsp}\uL + \tilde{m}_{u}\right)\bm{\delta}_{\tilde{\bm{h}}} / ||\bm{\delta}_{\tilde{\bm{h}}}||^{2}.
    \end{aligned}
\end{equation}
As for soft \ac{sfmds}, we compute $\bm{\delta}_{\tilde{\bm{h}}}$ using~\eqref{eq:proj-vector} with the outward-pointing normal $\bm{n}_{\tilde{\bm{h}}}\!=\!\nabla_{\bm{y}_{t}} d(\ye, \bm{y}_{t})^2$ of $\dXaLonly$. Since $\bm{J}_{\psi_{\theta}}$ is invertible, $\tilde{u}_{\theta}^{\Pi}$ is mapped to the original space using~\eqref{eq:u_relation}, yielding the hard-constrained flow matching dynamics
\begin{equation}
\label{eq:constrained_u}
    u^{\Pi}_{\theta} = \bm{J}_{\psi_{\theta}}^{-1} \tilde{u}^{\Pi}_{\theta}.
\end{equation}
As a result, the system $f_{\theta}$ is globally asymptotically stable.

We can further exploit the invertibility of $\bm{J}_{\psi_{\theta}}$ to control the convergence rate towards the equilibrium $\xe$ by imposing a chosen task-space margin $m_{\bm{h}}$. This is achieved by relating $m_{\bm{h}}$ with its latent counterpart $\tilde{m}_{\bm{h}}$ used to compute $\delta_{\tilde{\bm{h}}}$. 
Specifically, the margins relate as $\tilde{m}_{\bm{h}}\!=\!{m_{\bm{h}}\lVert\tilde{\bm{n}}_{\bm{h}}\lVert}/{\lVert\bm{n}_{\bm{h}}\lVert}$ with $\tilde{\bm{n}}_{\bm{h}}\!=\!\nabla_{\bm{y}_{t}} d(\ye, \bm{y}_{t})^2$ and $\bm{n}_{\bm{h}}\!=\!\bm{J}_{\psi_{\theta}}^{-1}\tilde{\bm{n}}_{\bm{h}}$. Hard SFMDS is summarized by Algorithm~\ref{algo:hard-sfmds}.
\begin{remark}
For soft \ac{sfmds}, a neural network parametrizes $u_\theta$ directly in task (velocity) space $\dot{\sX}$. For hard \ac{sfmds}, in contrast, a potentially identical, unconstrained, neural network outputs $\tilde{u}_\theta$ in $\dot{\sL}$, from which the constrained task-space dynamics are obtained via~\eqref{eq:constrained_u}, see Fig.~\ref{fig:method_2d_example}.
\end{remark}

\subsection{Discrete-Time Systems: the Forward-Euler Case}
\label{subsec:discrete-time-systems}
In the previous section, we enforced asymptotic stability in a \emph{continuous-time} flow matching dynamical system. In practice, however, such systems are implemented via numerical integration, yielding \emph{discrete-time} dynamics whose properties depend on the chosen integration scheme. In general, asymptotic stability in continuous time does not carry over to the discretized systems, so the discrete-time behavior must be explicitly considered.
%
In particular, we address the scenario where the system $f_{\theta}$ is evolved using \emph{forward Euler integration}, as shown in Fig.~\ref{fig:method_2d_example}, though the approach naturally extends to other integrators.
Readers primarily interested in the core principles of our method, rather than this extension for robust practical deployment, may skip the remainder of this section.

\paragraph{Discretizing $f_{\theta}$}
First, we analyze the discrete dynamics in the latent space $\sL$ where the constraints are imposed.
For the discrete-time dynamics $\bm{y}_{t+1} \!=\! \bm{y}_t + \dot{\bm{y}}_t\,\Delta t$, with $\Delta t >0$, the Lyapunov condition $\dot{V}^{\sL}(\bm{y}_t)<0$ becomes
\begin{equation}
    V^{\sL}(\bm{y}_{t+1}) =d(\bm{y}_{t+1}, \ye)^2 < d(\bm{y}_t, \ye)^2 = V^{\sL}(\bm{y}_t),
\end{equation}
thus altering the set of admissible velocities $\dXaLonly$. Substituting $\bm{y}_{t+1}$ and writing the distance as a norm yields $\lVert \bm{y}_t + \dot{\bm{y}}_t\,\Delta t - \ye \rVert < \lVert \bm{y}_t - \ye \rVert$.
Dividing both sides by $\Delta t$ and rearranging terms leads to
\begin{equation}
\label{eq:ball_h}
    \lVert \dot{\bm{y}}_t - \underbrace{(\ye - \bm{y}_t)/\Delta t}_{\bm{c_\bm{h}}} \rVert < \underbrace{\lVert \ye - \bm{y}_t \rVert/\Delta t}_{r_{\bm{h}}},
\end{equation}
defining the ball $\dXaLB = \{ \dot{\bm{y}}_t \in \dXL : \lVert \dot{\bm{y}}_t - \bm{c_\bm{h}} \rVert < r_{\bm{h}}\}$ of admissible velocities, with center $\bm{c_\bm{h}}$ and radius $r_{\bm{h}}$.

\begin{figure}[t]
    \centering
    \includegraphics[width=.8\linewidth,trim= 550 500 20 600,clip]{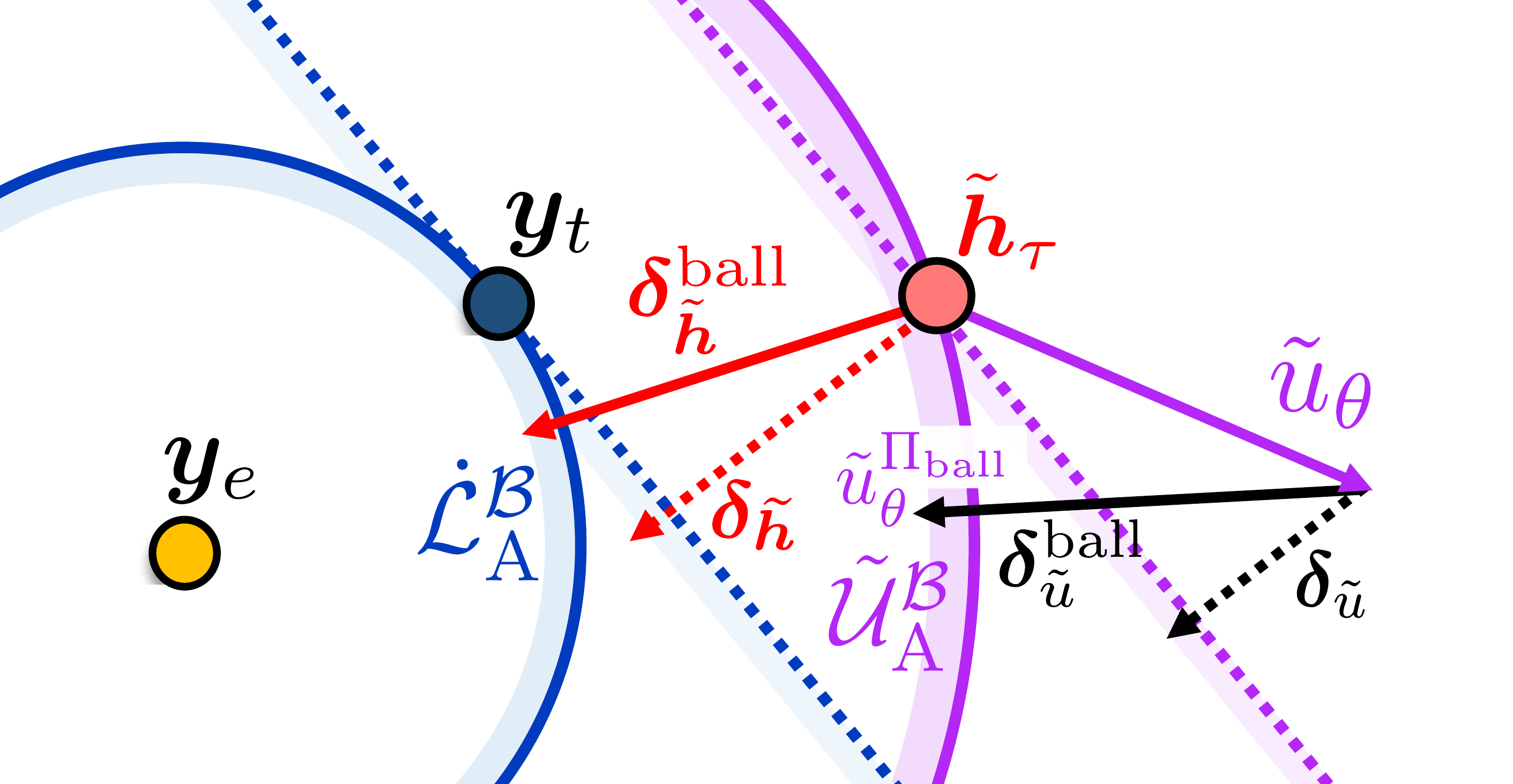}
    \caption{Overlap between $\sX$, $\dot{\sX}$, and $\mathcal{U}$. Dotted lines denote the admissible half-spaces, while solid lines denote the admissible balls, whose depiction illustrates their effect in $\sX$. The admissible sets in velocity space are shown in blue, and those in $\mathcal{U}$-space are shown in purple.}
    \label{fig:balls}
\end{figure}

\paragraph{Discretizing $u_{\theta}$}
Similarly, for the discrete latent flow matching dynamics 
$
\tilde{\bm{h}}_{\tau+1} \!=\! \tilde{\bm{h}}_{\tau} + \tilde{u}_{\theta}(\tilde{\bm{h}}_{\tau})\,\Delta\tau,
$
with ${\Delta\tau\!>\!0}$, LaSalle’s decrease condition~\eqref{eq:invariance_dxae} becomes
\begin{equation}
    {H_{\bm{h}}(\tilde{\bm{h}}_{\tau+1}) \!<\! H_{\bm{h}}(\tilde{\bm{h}}_{\tau})}.
\end{equation} 
Following \eqref{eq:H-func}, we define $H_{\bm{h}}$ as the squared distance between $\tilde{\bm{h}}_{\tau}$ and $\dXaLB$. We compute this distance as the norm of the vector $\bm{\delta}^{\text{ball}}_{\tilde{\bm{h}}}$ that projects $\tilde{\bm{h}}_{\tau}$ onto $\dXaLB$ (see Fig.~\ref{fig:balls}). Then, from this ball projection vector, LaSalle’s condition can be expressed as
\begin{equation}
\label{eq:lasalle_ball}
    \lVert \bm{\delta}^{\text{ball}}_{\tilde{\bm{h}}}(\tilde{\bm{h}}_{\tau+1}) \rVert < \lVert \bm{\delta}^{\text{ball}}_{\tilde{\bm{h}}}(\tilde{\bm{h}}_{\tau}) \rVert,
\end{equation}
which is the discrete-time equivalent of \eqref{eq:def-mB}. The norm of $\bm{\delta}^{\text{ball}}_{\tilde{\bm{h}}}$ admits the closed form
\begin{equation}
\label{eq:proj_La}
    \lVert \bm{\delta}^{\text{ball}}_{\tilde{\bm{h}}}(\tilde{\bm{h}}_{\tau}) \rVert 
    = \sigma_r( \lVert \tilde{\bm{h}}_{\tau} - \bm{c}_{\bm{h}} \rVert - (r_{\bm{h}}-b_{\bm{h}})).
\end{equation}
Note that the scalar $b_{\bm{h}}$ defining the margin $m_{\bm{h}}$ in~\eqref{eq:proj-vector} was integrated and subtracted from $r_{\bm{h}}$ to enforce the strict inequality~\eqref{eq:ball_h}. When $\tilde{\bm{h}}_{\tau}\notin\dXaL$, replacing~\eqref{eq:proj_La} and $\tilde{\bm{h}}_{\tau+1}$ in~\eqref{eq:lasalle_ball} leads to
\begin{equation}
\label{eq:u_ball_condition}
    \lVert \tilde{u}_{\theta} - \underbrace{(\bm{c}_{\bm{h}}-\tilde{\bm{h}}_{\tau}) / \Delta \tau}_{\bm{c}_{u}} \rVert < \underbrace{\lVert \bm{c}_{\bm{h}}-\tilde{\bm{h}}_{\tau} \rVert / \Delta \tau}_{r_u},
\end{equation}
resulting in the ball ${\dUaL \!=\! \{ \tilde{u}_\theta \in \tilde{\sU}: \lVert \tilde{u}_\theta - \bm{c}_{u} \rVert \!<\! r_{u} \}}$ of admissible velocities for $\tilde{u}_{\theta}$ (see Fig.~\ref{fig:balls}). 

\paragraph{Soft-constrained discrete-time $u_\theta$}
The admissible ball $\dUaL$ is contained within the continuous-time admissible half-space enforced in~\eqref{eq:stability_loss_sfmds}. We therefore augment~\eqref{eq:stability_loss_sfmds}, hereafter denoted by $\ell_{\text{half}}$, to also enforce the ball constraint as
\begin{equation}
    \ell_{\text{SFMDS}} = \ell_{\text{half}} + \gamma \ell_{\text{ball}},
\end{equation}
with $\gamma \in \euclideanspace_{>0}$.
To remain consistent with the half-space formulation, $\ell_{\text{ball}}$ penalizes violations of the half-space induced by the hyperplane tangent to the ball $\dUaL$ at the projection of $u_\theta$ onto this set. This projection is given by
\begin{equation}
    \label{eq:projection_ball_u}
    \begin{aligned}
        \tilde{u}_{\theta}^{\Pi_\text{ball}} &= \tilde{u}_{\theta} + \bm{\delta}_{\tilde{u}}^{\text{ball}}, \\
        \text{with } \bm{\delta}^{\text{ball}}_{\tilde{u}} &= \sigma_r\!\left( \lVert \bm{c}_{u} - \tilde{u}_{\theta} \rVert - (r_{u}-b_{u}) \right)
        \frac{\bm{c}_{u} - \tilde{u}_{\theta}}{\lVert \bm{c}_{u} - \tilde{u}_{\theta} \rVert},
    \end{aligned}
\end{equation}
with $\tilde{u}_{\theta}=\bm{J}_{\psi_{\theta}}u_{\theta}$ as in~\eqref{eq:stability_loss_sfmds}. Moreover, we have incorporated the offset $b_{u}$ to enforce convergence of $\tilde{\bm{h}}_\tau$ towards $\dXaLB$ by $\tau=1$. The boundary of this half-space has normal vector ${\bm{n}_{u} \!=\! -\bm{\delta}^{\text{ball}}_{\tilde{u}}}$. As this boundary does not pass through the origin, the hyperplane definition includes an additional margin ${m_{\tilde{u}}^{+} = (\bm{\delta}^{\text{ball}}_{\tilde{u}})^\trsp \tilde{u}_{\theta}^{\Pi_{\text{ball}}}}$. This yields the half-space condition $(-\bm{\delta}^{\text{ball}}_{\tilde{u}})^\trsp \tilde{u}_{\theta} + m_{\tilde{u}}^{+} \leq 0$, and therefore the penalty
\begin{equation}
    \ell_{\text{ball}} = \sigma_r\!\left( (-\bm{\delta}^{\text{ball}}_{\tilde{u}})^\trsp \tilde{u}_{\theta} + m_{\tilde{u}}^{+} \right).
\end{equation}

While $\ell_{\text{ball}}\!=\!0$ does not strictly guarantee feasibility, iterative minimization drives $\tilde{u}_\theta$ arbitrarily close to the admissible ball.

\paragraph{Hard-constrained discrete-time $u_\theta$}
The projected vector $\tilde{u}_{\theta}^{\Pi_{\text{ball}}}$~\eqref{eq:projection_ball_u} can be used directly in hard \ac{sfmds}, yielding
\begin{equation}
    u^{\Pi_{\text{ball}}}_{\theta} = \bm{J}_{\psi_{\theta}}^{-1} \tilde{u}^{\Pi_{\text{ball}}}_{\theta}.
\end{equation}
We note that a few number of edge cases may still arise from the discretization of $u_{\theta}$. They are discussed, together with proposed solutions, in App.~\ref{app:edge_cases_u}.

%% file: Algorithms/soft_SFMDS.tex
\newcommand{\algosection}[2][black]{%
  \BlankLine
  \noindent\textcolor{#1}{\hdashfill}\\[-0.4em]
  \textcolor{#1}{\tcp*[h]{#2}}\\[-0.4em]
  \noindent\textcolor{#1}{\hdashfill}
}
\newcommand{\hdashfill}{%
  \leavevmode
  \xleaders\hbox{-\,}\hfill\kern0pt
}

\definecolor{softil}{named}{MidnightBlue}
\definecolor{softpi}{named}{BrickRed}
\definecolor{softst}{named}{ForestGreen}
\definecolor{hardil}{named}{Plum}
\definecolor{hardvf}{named}{BlueViolet}

\begin{algorithm}[h!]
\small
\caption{Soft SFMDS}
\label{algo:soft-sfmds}

\SetKwFunction{ILloss}{il\_loss}
\SetKwFunction{PIloss}{pi\_loss}
\SetKwFunction{ASloss}{as\_loss}

\KwIn{Initial vector field $u_\theta$, encoder $\psi_\theta$, dataset $\mathcal{D}=\{(\bm{x}_t^{(i)}, \dot{\bm{x}}_t^{(i)})\}_{i=1}^N$, boundary $\partial \mathcal{X}$, latent margin $\tilde{\bm{m}}_h$, equilibrium $\xe$, scalar weights $\lambda_{\text{IL}}$, $\lambda_{\text{PI}}$, $\lambda_{\text{AS}}$. }
\KwOut{Trained stable vector field $u_\theta$}
\Repeat{convergence}{
 \textcolor{softil}{$\ell_{\text{IL}} \leftarrow$ \ILloss{$u_\theta, \mathcal{D}$}} \\
 \textcolor{softpi}{$\ell_{\text{PI}} \leftarrow$ \PIloss{$u_\theta$, $\partial \mathcal{X}$}} \\
 \textcolor{softst}{$\ell_{\text{AS}} \leftarrow$ \ASloss{$u_\theta$, $\psi_\theta$, $\tilde{\bm{m}}_h$, $\xe$}} \\
Compute $\ell_{\text{SFMDS}} = \lambda_{\text{IL}}\ell_{\text{IL}} + \lambda_{\text{PI}}\ell_{\text{PI}} +  \lambda_{\text{AS}}\ell_{\text{AS}}$.\\
Update $u_\theta$ and $\psi_\theta$ by minimizing $\ell_{\text{SFMDS}}$ w.r.t. $\theta$. \\
}

\algosection[softil]{Imitation Learning Loss\footnotemark}

\SetKwFunction{ILloss}{il\_loss}
\SetKwProg{Fn}{Function}{:}{}
\KwIn{Vector field $u_\theta$, dataset $\mathcal{D}=\{(\bm{x}_t^{(i)}, \dot{\bm{x}}_t^{(i)})\}_{i=1}^N$}
\KwOut{Imitation learning loss $\ell_{\text{IL}}$}
\Fn{\ILloss{$u_\theta, \mathcal{D}$}}{

Sample batch  $(\bm{x}_t, \dot{\bm{x}}_t) \sim \mathcal{D}$\\
Sample $\bm{h}_0 \sim \mathcal{N}(\bm{0}, \bm{I})$\\
Sample $s_\tau \sim \mathrm{Unif}([0,1])$\\
Set $\bm{h}_1 \leftarrow \dot{\bm{x}}_t$\\
Compute $\bm{h}_\tau \leftarrow \operatorname{GaussianPath}(\bm{h}_0, \bm{h}_1, s_\tau)$ \hfill  Eq.~\eqref{eq:stable_gauss_path}\\
Compute target velocity $\bm{u} \leftarrow {\lambda_h}(\bm{h}_1 - \bm{h}_\tau)$\\
\Return $\ell_{\text{IL}}(u_\theta(\bm{h}_\tau, s_\tau; \bm{x}_t), \bm{u})$\hfill Eq.~\eqref{eq:il} 
}

\algosection[softpi]{Positive Invariance Loss}

\SetKwFunction{PIloss}{pi\_loss}
\SetKwProg{Fn}{Function}{:}{}
\KwIn{Vector field $u_\theta$, boundary $\partial \mathcal{X}$}
\KwOut{Positive invariance loss $\ell_{\text{PI}}$}
\Fn{\PIloss{$u_\theta$, $\partial \mathcal{X}$}}{

Sample $\bm{h}_\tau \sim \mathrm{Unif}(\dot{\mathcal{X}})$\\
Sample $s_\tau \sim \mathrm{Unif}([0,1])$\\
Sample boundary states $\bm{x}_t \sim \mathrm{Unif}(\partial \mathcal{X})$\\
\tcp{Projection $\bm{h}_\tau$ onto $\dot{\mathcal{X}}_{\text{A}}$}
Compute normal $\bm{n}_h \leftarrow \operatorname{OutwardNormal}(\bm{x}_t, \partial \mathcal{X})$\\
Set $\bm{m}_h \leftarrow \bm{0}$ \tcp{non-strict inequality}
Compute $\bm{\delta}_h \leftarrow \operatorname{Proj}(\bm{h}_\tau, \bm{n}_h, \bm{m}_h)$ \hfill Eq.~\eqref{eq:proj-vector} \\ 
\tcp{Penalize $u_\theta$ according to $\mathcal{U}_{\text{A}}$}
Compute margin $\bm{m}_u \leftarrow {\lambda_h} \lVert \bm{\delta}_h \rVert^2$\\
Compute normal vector $\bm{n}_u \leftarrow -\bm{\delta}_h$\\
\Return $\ell_{\text{PI}}(u_\theta(\bm{h}_\tau, s_\tau; \bm{x}_t), \bm{n}_u, \bm{m}_u)$ \hfill Eq.~\eqref{eq:stability_loss}
}
\algosection[softst]{Asymptotic Stability Loss}

\SetKwFunction{ASloss}{as\_loss}
\SetKwProg{Fn}{Function}{:}{}
\KwIn{Vector field $u_\theta$, encoder $\psi_\theta$, latent margin $\tilde{\bm{m}}_h$, equilibrium $\xe$}
\KwOut{Asymptotic stability loss $\ell_{\text{AS}}$ }
\Fn{\ASloss{$u_\theta$, $\psi_\theta$, $\tilde{\bm{m}}_h$, $\xe$}}{
Sample $\bm{h}_\tau \sim \mathrm{Unif}(\dot{\mathcal{X}})$\\
Sample $s_\tau \sim \mathrm{Unif}([0,1])$\\
Sample states $\bm{x}_t \sim \mathrm{Unif}(\mathcal{X})$\\
\tcp{Mappings to latent space}
Compute latent FM states $\tilde{\bm{h}}_\tau \leftarrow \bm{J}_{\psi_\theta}(\bm{x}_t)\,\bm{h}_\tau$\\
Compute latent FM velocity $\tilde{\bm{u}}_\theta \leftarrow \bm{J}_{\psi_\theta}(\bm{x}_t)\,u_\theta(\bm{h}_\tau, s_\tau; \bm{x}_t)$\\
Compute latent states $\bm{y}_t, \ye \leftarrow \psi_\theta(\bm{x}_t), \psi_\theta(\xe)$\\
\tcp{Compute projection onto latent $\dot{\mathcal{L}}_{\text{A}}$}
Compute normal $\tilde{\bm{n}}_h \leftarrow \nabla_{\bm{y}}\, V^{\sL}$ \hfill From Eq.~\eqref{eq:latent_ad_set}\\
Compute $\tilde{\bm{\delta}}_h \leftarrow \text{Proj}(\tilde{\bm{h}}_\tau, \tilde{\bm{n}}_h, \tilde{\bm{m}}_h)$ \hfill Eq.~\eqref{eq:proj-vector}\\
\tcp{Penalize $\tilde{\bm{u}}_\theta$ according to latent $\tilde{\mathcal{U}}_{\text{A}}$}
Compute margin $\tilde{\bm{m}}_u \leftarrow {\lambda_h}\,\lVert \tilde{\bm{\delta}}_h \rVert^2$\\
Compute normal vector $\tilde{\bm{n}}_u \leftarrow -\tilde{\bm{\delta}}_h$\\
\Return $\ell_{\text{AS}}(\tilde{\bm{u}}_\theta, \tilde{\bm{n}}_u, \tilde{\bm{m}}_u)$ \hfill Eq.~\eqref{eq:stability_loss_sfmds}
}
\end{algorithm}

%% file: Algorithms/hard_SFMDS.tex
\begin{algorithm}[h]
\small
\caption{Hard SFMDS}
\label{algo:hard-sfmds}
\SetKwFunction{HardILloss}{hard\_il\_loss}
\KwIn{Initial latent vector field $\tilde{u}_\theta$, encoder $\psi_\theta$, dataset $\mathcal{D}=\{(\bm{x}_t^{(i)}, \dot{\bm{x}}_t^{(i)})\}_{i=1}^N$, latent margin $\tilde{m}_{\bm{h}}$, equilibrium $\xe$}
\KwOut{Trained stable vector field $u_\theta$}
\Repeat{convergence}{
\textcolor{hardil}{$\ell_{\text{IL}} \leftarrow$ \HardILloss{$\tilde{u}_\theta, \mathcal{D}, \psi_{\theta}, \tilde{m}_{\bm{h}}$}} \\
Update $\tilde{u}_\theta$ and $\psi_\theta$ by minimizing $\ell_{\text{IL}}$ w.r.t. $\theta$. \\
}

\algosection[hardil]{Hard Imitation Learning Loss }

\SetKwFunction{HardILloss}{hard\_il\_loss}
\SetKwFunction{ForwardPass}{hard\_forward}
\SetKwProg{Fn}{Function}{:}{}
\KwIn{Vector field $u_\theta$, dataset $\mathcal{D}=\{(\bm{x}_t^{(i)}, \dot{\bm{x}}_t^{(i)})\}_{i=1}^N$, encoder $\psi_{\theta}$, latent margin $\tilde{m}_{\bm{h}}$}
\KwOut{Imitation learning loss $\ell_{\text{IL}}$}
\Fn{\HardILloss{$u_\theta, \mathcal{D}, \psi_{\theta}, \tilde{m}_{\bm{h}}$}}{

Sample batch  $(\bm{x}_t, \dot{\bm{x}}_t) \sim \mathcal{D}$\\
Sample $\bm{h}_0 \sim \mathcal{N}(\bm{0}, \bm{I})$\\
Sample $s_\tau \sim \mathrm{Unif}([0,1])$\\
Set $\bm{h}_1 \leftarrow \dot{\bm{x}}_t$\\
Compute $\bm{h}_\tau \leftarrow \operatorname{GaussianPath}(\bm{h}_0, \bm{h}_1, s_\tau)$ \hfill  Eq.~\eqref{eq:stable_gauss_path}\\
Compute target velocity $\bm{u} \leftarrow {\lambda_h}(\bm{h}_1 - \bm{h}_\tau)$\\
Compute \textcolor{hardvf}{$u^{\Pi}_{\theta}\leftarrow$\ForwardPass{$u_\theta, \psi_{\theta}, \bm{h}_{\tau}, s_\tau, \bm{x}_t, \tilde{m}_{\bm{h}}$}}\\
\Return $\ell_{\text{IL}}(u^{\Pi}_{\theta}, \bm{u})$\hfill Eq.~\eqref{eq:il}
}
\algosection[hardvf]{Constrained Vector Field}

\SetKwFunction{ForwardPass}{hard\_forward}
\SetKwProg{Fn}{Function}{:}{}
\KwOut{Constrained inference $u^{\Pi}_{\theta}$}
\Fn{\ForwardPass{$\tilde{u}_\theta, \psi_{\theta}, \bm{h}_{\tau}, s_\tau, \bm{x}_t, \tilde{m}_{\bm{h}}$}}{
Compute latent FM states $\tilde{\bm{h}}_\tau \leftarrow \bm{J}_{\psi_\theta}(\bm{x}_t)\,\bm{h}_\tau$\\
Compute latent states $\bm{y}_t, \ye \leftarrow \psi_\theta(\bm{x}_t), \psi_\theta(\xe)$\\
\tcp{Compute projection onto latent $\dot{\mathcal{L}}_{\text{A}}$}
Compute normal $\tilde{\bm{n}}_h \leftarrow \nabla_{\bm{y}}\, V^{\sL}$ \hfill From Eq.~\eqref{eq:latent_ad_set}\\
Compute $\tilde{\bm{\delta}}_h \leftarrow \operatorname{Proj}_{h}(\tilde{\bm{h}}_\tau, \tilde{\bm{n}}_h, \tilde{m}_{\bm{h}})$ \hfill Eq.~\eqref{eq:proj-vector}\\

Compute margin $\tilde{m}_u \leftarrow {\lambda_h}\,\lVert \tilde{\bm{\delta}}_h \rVert^2$\\
Compute normal vector $\tilde{\bm{n}}_u \leftarrow -\tilde{\bm{\delta}}_h$\\
Compute $\delta_{\tilde{u}}\leftarrow \operatorname{Proj}_{u}(\tilde{u}_\theta, \tilde{\bm{n}}_u, \tilde{m}_u)$ \hfill Eq.~\eqref{eq:projection_hard_sfmds}\\
Get projection $\tilde{u}_{\theta}^{\Pi} \leftarrow \tilde{u}_{\theta} + \delta_{\tilde{u}}$ \\ 
Map to task space $u^{\Pi}_{\theta} \leftarrow \bm{J}_{\psi_{\theta}}^{-1} \tilde{u}^{\Pi}_{\theta}$

\Return $u^{\Pi}_{\theta}$
}
\end{algorithm}

%% file: Sections/07_Method_C.tex
\section{Stable Flow Matching Dynamical Systems on Lie Groups}
\label{sec:sfmds_lie} 

We extend \ac{sfmds} to Lie groups, allowing us to encode, e.g., the orientation part or the pose of end-effector motions.
We define a flow matching dynamical system on a Lie group $\mathcal{X}$ as the output of a flow ${\phi_{\theta}:\mathfrak{g}\times[0,1]\times\mathcal{X} \to \mathfrak{g}}$ on the Lie algebra of $\sX$, represented in a global frame at the identity via the inverse adjoint operator $\textrm{Ad}_{\bm{x}}^{-1}$ as
\begin{align}
\label{eq:u_lie_integration}
\dot{\bm{x}}_t &= \textrm{Ad}_{\bm{x}}^{-1} f_{\theta}(\bm{x}_{t}, \bm{\omega}_t) = \textrm{Ad}_{\bm{x}}^{-1}\phi_{\theta}(\bm{\omega}_t, s_0=0,\tau=1;\bm{x}_{t}) \nonumber \\&= \textrm{Ad}_{\bm{x}}^{-1}\bm{\omega}_t + \int_{0}^{1}\textrm{Ad}_{\bm{x}}^{-1}u_{\theta}(\bm{h}_{\tau}, s_\tau;\bm{x}_{t})d\tau,
\end{align}
where the last equality follows from~\eqref{eq:stochastic_dynsym} and the linearity of $\textrm{Ad}_{\bm{x}}^{-1}$. Similarly as in the Euclidean case, we learn a vector field $u_{\theta}$ that optimizes~\eqref{eqn:problem_formulation} subject to $ \textrm{Ad}_{\bm{x}}^{-1}{f_\theta(\bm{x}_t, \bm{\omega}_t) \in \admfuncset_\text{A}}$.

\subsection{Soft \ac{sfmds} on Lie Groups}
For soft \ac{sfmds}, asymptotic stability is enforced analogously to the Euclidean case via the loss~\eqref{eq:stability_loss_sfmds}. The key distinction is that the vector field $u_\theta\in\mathfrak{g}$ is, following~\eqref{eq:u_lie_integration}, constrained by evaluating the stability losses on $\textrm{Ad}_{\bm{x}}^{-1}u_{\theta}$. 
Note that the positive invariance loss from~\eqref{eq:positive_invariant_set} is not necessary for closed Lie groups such as $\sS^1$, $\sS^3$, and $\mathrm{SO}(3)$. 

\subsection{Hard \ac{sfmds} on Lie Groups}
Regarding hard \ac{sfmds}, we recall that enforcing global asymptotic stability through hard constraints requires a diffeomorphic mapping $\psi_{\theta}$ between $\sX$ and $\sL$, where both spaces are now Lie groups. As in prior works~\cite{Urain22:StableSE3, mohammadi2023neural}, we use the logarithmic map (see App.~\ref{app:lie_groups} for details) to map elements from the Lie group $\sX$ to elements $\bm{\xi}_{t} = \text{log}(\bm{x}_{t})$ on its Lie algebra $\mathfrak{g}_{\sX}$. We then define the Euclidean diffeomorphic network $\psi_{\theta}:\mathfrak{g}_{\sX}\to\mathfrak{g}_{\sL}$ between task-space and latent-space Lie algebras. Note that special care must must be taken to handle wrapping effects, e.g., in $\textrm{SO}(3)$, as a $360^\circ$ rotation corresponds to the same element in $\sX$ as a $0^\circ$ rotation. In such case, the logarithmic map is multivalued, violating the required bijectivity of $\psi_{\theta}$. Inspired by~\cite{mohammadi2023neural,Urain22:StableSE3}, we enforce the logarithmic map to remain diffeomorphic by restricting $\psi_{\theta}$ to the region of the Lie algebra where wrapping does not occur, known as the \emph{first cover}. Effectively, this leads to the first cover of $\mathfrak{g}_{\sX}$ being a positively invariant set w.r.t. $f_{\theta}$. We proceed to describe how this constraint is addressed for the Lie groups considered in this work.

\subsubsection{Torus $\mathbb{T}^2$}
For the 2-torus, the first cover corresponds to the region in which each angular coordinate is restricted to the interval $\theta_{i,t} \in (-\pi, \pi)$. Hence, we can construct the mapping
\begin{equation}
\bm{\xi}_t = \pi \tanh\big(\psi_\theta^{-1}\, (\bm{y}_t)\big),
\end{equation}
which guarantees that each angular component remains within its principal interval. Since this mapping is invertible, we obtain the following inverse
\begin{equation}
\bm{y}_t = \psi_\theta\left( \operatorname{arctanh}\left(\frac{\bm{\xi}_t}{\pi}\right)\right)
\end{equation}
from which we obtain a map from the first cover of $\mathfrak{g}_{\sX}$ to the latent space that prevents wrapping.

\subsubsection{Special Euclidean group $\mathrm{SE}(3)$}
For the special Euclidean group, the Lie algebra representation of the rotation component $\bm{\xi}^r_t \in \mathfrak{so}(3)$ must satisfy $\lVert \bm{\xi^r}_t \rVert < \pi$ for $\bm{\xi}_t$ to remain within the first cover. We refer to this region as the $\pi$-ball $B_\pi$.
Let $\psi'_{\theta}: {\mathfrak{so}(3)}_{\mathcal{X}} \to {\mathfrak{so}(3)}_{\mathcal{L}}$ denote the diffeomorphic mapping of the rotation component. Following~\cite{mohammadi2023neural}, we ensure that the mapping remains within $B_\pi$ by first applying a unit-box constraint on the output, analogous to the approach used for the torus,
\begin{equation}
{\widetilde{\bm{\xi}^r_t}} = \pi\tanh\big({\psi'_\theta}^{-1}( \, \bm{y}^r_t)\big).
\end{equation}
We then smoothly map this unit cube to $B_\pi$ by scaling ${\widetilde{\bm{\xi}^r_t}}$ according to the distance $d$ from the origin of the latent space to the cube boundary along the direction $\mathbf{n}= \frac{{\widetilde{\bm{\xi}^r_t}}}{\Vert {\widetilde{\bm{\xi}^r_t}} \Vert}$, i.e., 
\begin{equation}
    d = \frac{1}{\Vert \mathbf{n}\Vert_\infty}, \quad {{\bm{\xi}^r_t}}=\frac{\widetilde{\bm{\xi}^r_t}}{d},
\end{equation}
thus ensuring that ${\bm{\xi}}^r_t \in B_\pi$, and therefore that ${\bm{\xi}}_t$ belongs to the first cover of $\mathfrak{g}_{\sX}$.
The inverse mapping is then expressed as
\begin{equation}
{{\bm{y}^r_t}} = \psi'_\theta\left(\operatorname{arctanh}\left(\frac{d}{\pi}{\bm{\xi} }^r_t\right)\right).
\end{equation}

%% file: Sections/08_Experiments.tex
\section{Experiments}
\label{sec:experiments}
We evaluate \ac{sfmds} across six experiments spanning benchmark learning-from-demonstration problems, complex action distributions, systems on Lie groups, real robotic manipulation tasks, and high-dimensional dynamical systems.
We consider: (1) LASA dataset~\citep{khansari-zadeh2011learning}, (2) LASA on $\mathbb{T}^2$ as a new benchmark dataset, (3) a novel multimodal dataset in $\mathbb{R}^2$, (4) the multimodal simulated pouring task on $\mathrm{SE}(3)$ from~\citep{lee2024mmp++}, (5) a real robotic insertion task on $\mathrm{SE}(3)$, and (6) a $1682$-dimensional reaction--diffusion system. We assess \emph{trajectory accuracy} using multiple distance-based metrics and \emph{asymptotic stability} by testing convergence from out-of-distribution initial conditions. As part of the baselines, we consider a behavior cloning flow-matching dynamical system variant (BC$_{\text{FMDS}}$), which uses the same network architecture and training procedure as the soft \ac{sfmds} variant but without the stability losses. Due to its unconstrained nature, it serves as a reference for achievable accuracy and, crucially, highlights the lack of asymptotic stability in unconstrained networks. Additional baselines are described for each individual experiment.

\textbf{Accuracy metrics.} We employ multiple distance metrics to quantify the accuracy of the \ac{sfmds} variants and considered baselines. For unimodal datasets, we report the Root Mean Squared Error (RMSE), Dynamic Time Warping Distance (DTWD)~\cite{muller2007dynamic}, and Fréchet Distance (FD)~\cite{eiter1994computing} between the demonstrations and the trajectories generated by the learned dynamics $f_{\theta}$. These metrics are computed by initializing $f_{\theta}$ from the same initial conditions as in the dataset, rolling out the dynamics for the same number of time steps, and comparing the resulting trajectories to the corresponding demonstrations, following standard practice~\citep{Rana2020:EuclideanizingFlows,Urain20:ImitationFlow,perez2024puma}.
For multimodal datasets, the aforementioned metrics are ill-defined due to the demonstrations' multimodal nature. Instead, we compare the set of rollouts produced by $f_{\theta}$ to the set of demonstrations using the Chamfer distance~\citep{barrow1977parametric}. Given the set $G$, consisting of all the points in the trajectories generated from $f_{\theta}$, and the corresponding demonstration set $D$, the Chamfer distance is defined as
$\mathrm{CH}(G,D) := \frac{1}{|G|} \sum_{g \in G} \min_{d \in D} d(\bm{x}^g, \bm{x}^d)$.
We report both $\mathrm{CH}(G,D)$ and $\mathrm{CH}(D,G)$, as the former measures motion fidelity and the latter reflects the coverage of the demonstrations by the generated trajectories, i.e., the ability to capture multimodality. Multiple rollouts are sampled per initial condition to compute these metrics.

\subsection{Benchmark Datasets in $\euclideanspace^2$ and $\mathbb{T}^2$}
\label{subsec:exp_benchmarks}
We evaluate the performance of \ac{sfmds} and compare it with multiple state-of-the-art methods on three benchmark datasets: (1) LASA dataset~\citep{khansari-zadeh2011learning}, (2) LASA on $\mathbb{T}^2$ as a new benchmark dataset, (3) a novel multimodal dataset in $\mathbb{R}^2$. 
We sample $\bm{\omega}_{t}$ at every time step, except on the LASA dataset where we additionally consider the \emph{1S} variant which samples $\bm{\omega}_t$ only at $t=0$, as described in Sec.~\ref{subsec:asymptstable_FMDS}.
For the stability analysis, we simulated $900$ trajectories per model to compute the percentage of unsuccessful trajectories.

\input{Tables/unimodal_results}

\begin{figure}[t]
    \centering
    \includegraphics[width=0.9\linewidth,trim=0 0 0 7,clip]{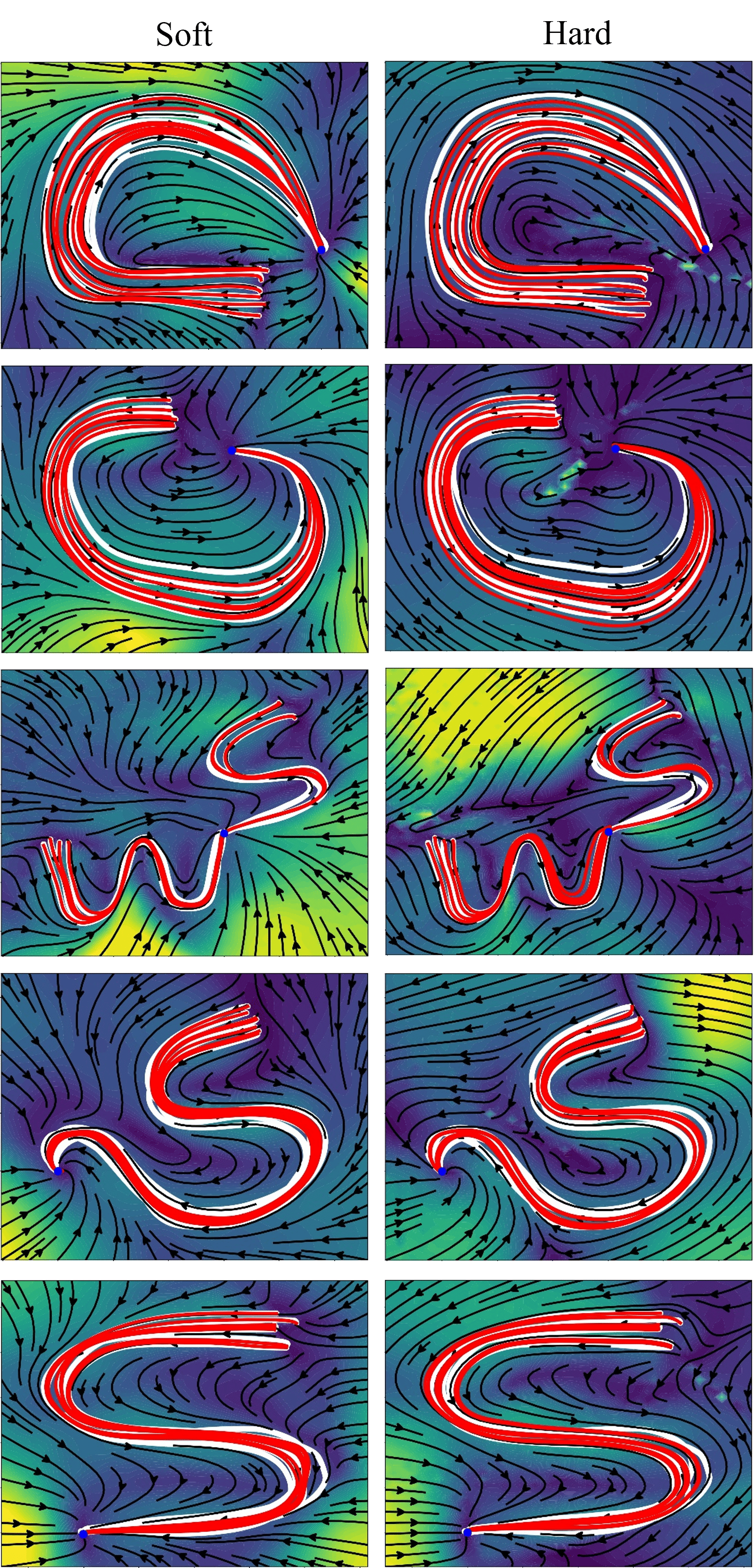}
    \caption{Vector fields of learned soft (left) and hard (right) \ac{sfmds} on LASA. White trajectories denote demonstrations, red trajectories show rollouts from the same initial conditions, and background color indicates normalized speed (lighter colors indicate higher values).}
    \label{fig:app_quali_lasa}
\end{figure}

\subsubsection{LASA}
\label{subsubsec:lasa_exps_main}
The LASA dataset~\citep{khansari-zadeh2011learning} is composed of $30$ two-dimensional human handwriting motions. Each motion, captured using a tablet PC, includes $7$ demonstrations of a desired trajectory from different initial positions. The demonstrated trajectories exhibit convergent behavior toward an attractor, so the dataset is well-suited for assessing the stability of learned dynamical systems.
We compare \ac{sfmds} with five benchmark methods: (1) CLF-DM~\citep{khansari-zadeh2014learning}, (2) Euclideanzing flows (Euc. Flows)~\citep{Rana2020:EuclideanizingFlows}, (3) a variant of Euclideanzing flows hereinafter referred to as Euc. Flows (NODE), where RealNVP is replaced by a Neural ODE (NODE)~\citep{chen2018neural}, (4) PUMA~\citep{perez2024puma}, and (5) Behavior Cloning (BC$_{\text{FMDS}}$). CLF-DM is a relatively early method constrained to stricter Lyapunov functions and is therefore expected to exhibit the lowest accuracy among the compared approaches. Euclideanzing flows employs latent Lyapunov functions together with RealNVP~\citep{dinh2017density} as an invertible mapping. Euc. Flows (NODE) follows the same approach as Euc. Flows, but uses Neural ODE (NODE)~\citep{chen2018neural}, known to be more expressive, as invertible mapping. PUMA also enforces stability using latent Lyapunov functions, but softly through a loss function. Finally, BC$_{\text{FMDS}}$ denotes an unconstrained flow matching dynamical system, trained without stability losses or constraints. All baselines, except BC$_{\text{FMDS}}$, feature unimodal behaviors. 

\begin{figure}[t]
    \centering
    \includegraphics[width=0.9\linewidth,trim=5 8 5 7,clip]{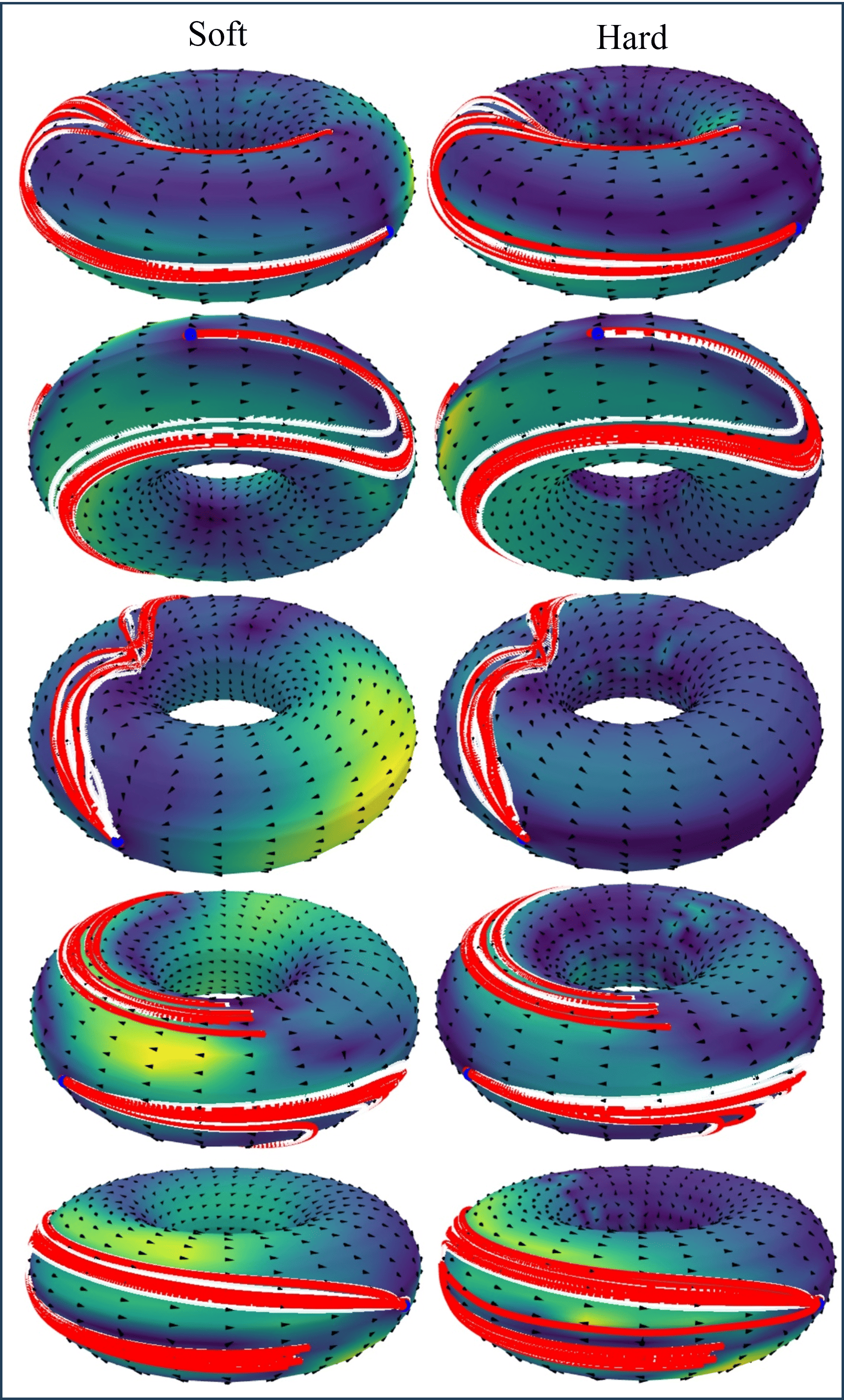}
    \caption{Vector fields of learned soft (left) and hard (right) \ac{sfmds} on LASA $\torus^{2}$. White trajectories are demonstrations, red trajectories show rollouts from the same initial conditions, and background color indicates normalized speed.}
    \label{fig:app_quali_lasa_t2}
    \vspace{-0.2cm}
\end{figure}
Figure~\ref{fig:app_quali_lasa} displays examples of learned soft and hard \ac{sfmds}, showing that both variants learn accurate and smooth behaviors. The effect of the positive invariance loss in soft \ac{sfmds} is evident, as the vector field at the boundary points inward. This contrasts with hard \ac{sfmds}, where global asymptotic stability does not require this behavior. 
As shown in Table~\ref{tab:lasa_ablation_results}, all methods achieve similar accuracy for the moderately complex motions of LASA. Moreover, \ac{sfmds} achieves competitive accuracy performance with the best-performing baselines. As expected, BC$_{\text{FMDS}}$ does not enforce stability, and $40\%$ of its trajectories fail to converge to the equilibrium. In contrast, all \ac{sfmds} variants consistently converge, with the sole exception of hard \ac{sfmds} \emph{1S} using RealNVP, where a negligible fraction of trajectories ($0.044\%$) does not exactly reach the goal. As described in App.~\ref{app:edge_cases_u}, this behavior might arise in edge cases where the Jacobian $\bm{J}_{\psi_\theta}$ varies too sharply due to overfitting, making accurate simulation of the dynamics challenging, potentially leading to spurious behavior. RealNVP has been shown to be particularly prone to such effects~\citep[Sec.~V.A]{perez2023stable}. 
We observe that the \emph{1S} and non-\emph{1S} \ac{sfmds} variants show comparable performance, motivating the use of the latter, more common in practice, for the remainder of the experiments. 

Regarding the baselines, Euc. Flows suffers from issues similar to those observed for hard \ac{sfmds} \emph{1S} with RealNVP and therefore also exhibits a few unsuccessful trajectories. Finally, CLF-DM constrains continuous-time dynamics while disregarding their discrete-time behavior, as elaborated in Sec.~\ref{subsec:discrete-time-systems}, which likewise leads to a small number of unsuccessful trajectories.

\subsubsection{LASA $\torus^2$}
 We introduce the LASA $\mathbb{T}^2$ dataset, which follow the same structure as the original LASA dataset and is constructed by projecting the demonstrations onto a two-dimensional unit torus $\mathbb{T}^2=\mathcal{S}^1 \times \mathcal{S}^1$. 
 Figure~\ref{fig:app_quali_lasa_t2} displays examples of \ac{sfmds} on $\torus^2$, showing that both \ac{sfmds} variants accurately capture trajectories that span a large portion of the manifold. 
 We compare \ac{sfmds} with baseline dynamical systems imposing asymptotic stability on Lie groups, namely LieImFlow~\citep{Urain22:StableSE3} and PUMA~\citep{perez2024puma}. LieImFlow extends the paradigm of latent Lyapunov–based methods with invertible networks, specifically employing NODE~\citep{chen2018neural}, to Lie groups, similarly to \ac{sfmds} in Sec.~\ref{sec:sfmds_lie}. From Table~\ref{tab:lasa_ablation_results}, we observe that all benchmarks lead to a similar accuracy, while BC$_{\text{FMDS}}$ fails to learn stable dynamics, leading to $32\%$ of failure cases.

\subsubsection{Multimodal $\euclideanspace^2$}
\label{sec:exp_multimodal_2d}
\begin{figure*}[t]
    \centering
    \includegraphics[width=\linewidth]{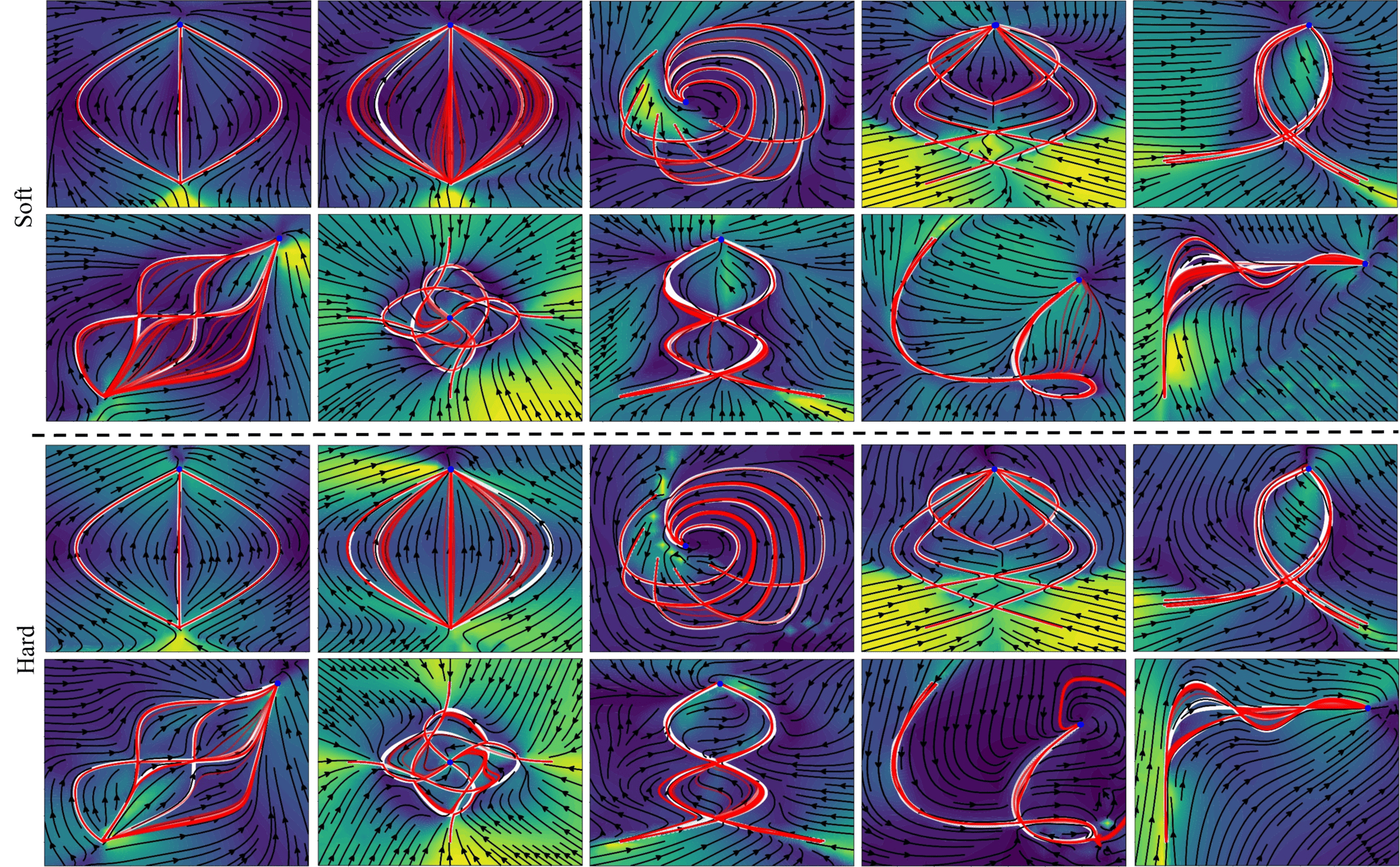}
    \caption{Vector fields of learned soft (top) and hard (bottom) \ac{sfmds} in the Multimodal $\euclideanspace^2$ dataset. White trajectories denote demonstrations, red trajectories show rollouts from the same initial conditions, and background color indicates normalized speed.}
    \label{fig:app_quali_multimodal}
\end{figure*}
We introduce the novel Multimodal $\mathbb{R}^2$ dataset, created to provide a multimodal benchmark for dynamical systems serving a role similar to that of LASA. Multimodal $\mathbb{R}^2$ is composed of $10$ different motions, each exhibiting between $3$ and $6$ intersecting trajectories. The complete dataset, together with the learned soft and hard \ac{sfmds}, is depicted in Fig.~\ref{fig:app_quali_multimodal}.
We observe that both \ac{sfmds} variants successfully capture the trajectory distributions present in the dataset, while generating asymptotically stable trajectories.
Note that, in some examples, not all modalities are perfectly captured, while in few cases certain sampled trajectories deviate from the data distribution. For example, for the second-to-last motion learned by hard \ac{sfmds}, at an intermediate stage of the trajectory, some samples diverge from the data support. However, they eventually reach the equilibrium, as expected given the method's stability guarantees. This is due to the substantial multimodality and sharp velocity transitions present in the introduced dataset, which are inherently difficult to learn. 

\input{Tables/multimodal_results}
We evaluate the accuracy performance of \ac{sfmds} by sampling $30$ trajectories for every initial condition, and compare them to the ground-truth trajectories using the Chamfer distance. Table~\ref{tab:multimodal_ablation_results} reports the accuracy and stability measure for both \ac{sfmds} variants, as well as for the Eucl. flow and PUMA baselines. 
In contrast to the previous unimodal datasets in $\euclideanspace^2$ and $\mathbb{T}^2$ where RealNVP and NODE led to similar performance for hard \ac{sfmds}, RealNVP failed to robustly learn the multimodal behaviors in this setting. 
Consequently, we use only NODE for the mapping $\psi_{\theta}$. All subsequent experiments involving hard \ac{sfmds} are also conducted exclusively with NODEs, as the tasks become increasingly challenging.
Table~\ref{tab:multimodal_ablation_results} shows that both \ac{sfmds} variants significantly outperform the state-of-the-art baselines in terms of accuracy on the multimodal dataset, which is expected given their unimodal nature. Lastly, \ac{sfmds} consistently exhibit asymptotically stable behavior. 

\subsection{Simulated Water Pouring on $\mathrm{SE}(3)$}
The $\mathrm{SE}(3)$ pouring dataset~\cite{lee2024mmp++} contains $5$ demonstrations for each of $2$ pouring tasks, labeled \emph{water} and \emph{wine}. 
We modified the dataset so that all demonstrations share the same equilibrium by selecting a single goal pose and interpolating the last $120$ steps of each trajectory in $\mathrm{SE}(3)$.

\begin{figure}[t]
    \captionsetup[subfloat]{labelformat=empty}
    \centering
    \raggedright
    \begin{minipage}{0.63\linewidth}
        \raggedright
        \subfloat[]{
              \begin{minipage}[c]{0.1\textwidth}
                \rotatebox{90}{\scriptsize SFMDS (Soft)}
              \end{minipage}%
              \hfill
              \begin{minipage}[c]{0.9\textwidth}
                \includegraphics[width=0.49\linewidth, trim=0 30 80 30,clip]{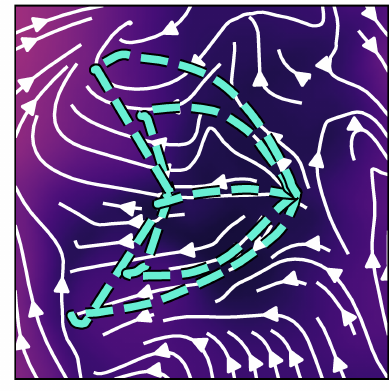}\hspace{-0.28em}
                \includegraphics[width=0.49\linewidth, trim=0 30 80 30,clip]{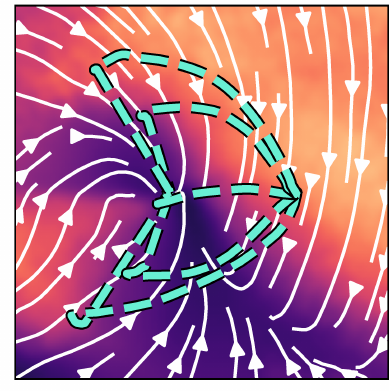}
              \end{minipage}
        }\\
        \vspace{-20pt}
        \subfloat[]{
              \begin{minipage}[c]{0.1\textwidth}
                \rotatebox{90}{\scriptsize SFMDS (Hard)}
              \end{minipage}%
              \hfill
              \begin{minipage}[c]{0.9\textwidth}
                \includegraphics[width=0.49\linewidth, trim=0 30 80 30,clip]{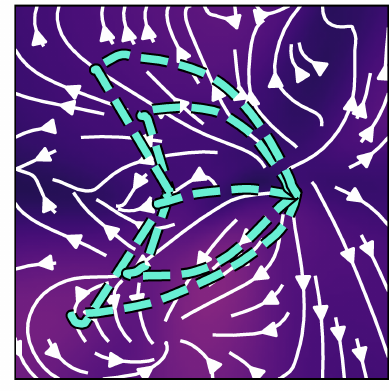}%
                \includegraphics[width=0.49\linewidth, trim=0 30 80 30,clip]{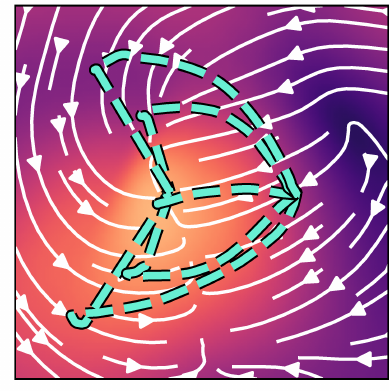}
              \end{minipage}
        }
    \end{minipage}%
    \begin{minipage}{0.3\linewidth}
        \raggedright
        \raisebox{0.5cm}{
            \includegraphics[width=\linewidth, trim=660 0 570 0,clip]{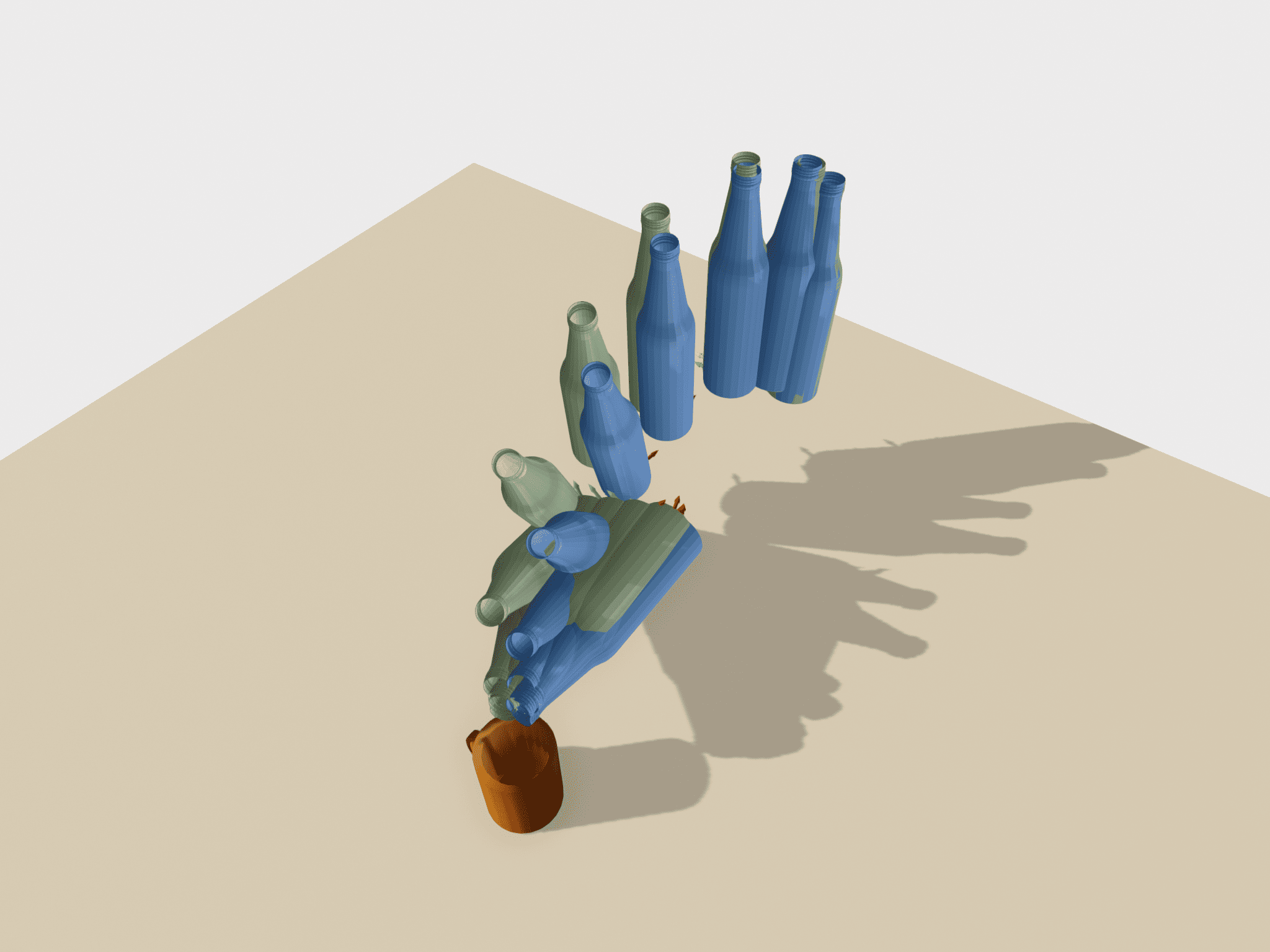}
        }
    \end{minipage}
    \begin{minipage}{0.63\linewidth}
     \vspace{-0.4cm}
       \centering \scriptsize \hspace{0.15\linewidth}(a) Init. Orientation\hspace{0.05\linewidth} (b) Final Orientation
    \end{minipage}
    \begin{minipage}{0.3\linewidth}
     \vspace{-0.4cm}
       \centering \scriptsize (c) Traj. $\mathrm{SE}(3)$ 
    \end{minipage}
    \vspace{-0.4cm}
    \caption{Learned \ac{sfmds} on the water-pouring task in $\mathrm{SE}(3)$. (a)-(b) Flow lines computed at constant orientation and height (white) and demonstrations (cyan) projected onto the $xy$-plane. Flow lines are initially multimodal (a) and converge at the end (b). Background color indicates the desired cup rotation (low/dark initially, higher/brighter near pouring). (c) Example demonstration (light bottles) and reproduced \ac{sfmds} trajectory (dark bottles).}
    \label{fig:water_qualitative}
\end{figure}

Figures~\ref{fig:water_qualitative}(a)-(b) depicts the learned \ac{sfmds} vector fields in a slice of $\mathrm{SE}(3)$, showing that both variants learn smooth and converging dynamics. Figure~\ref{fig:water_qualitative}(c) displays a demonstrated and a generated trajectory, showcasing that \ac{sfmds} stably encodes the demonstrated behaviors. 
Similarly to Sec.~\ref{sec:exp_multimodal_2d}, we compute the accuracy metrics for each trained model by generating $30$ rollouts and comparing them to the ground-truth trajectories using the Chamfer distance.
Table~\ref{tab:multimodal_ablation_results} shows that soft \ac{sfmds} outperforms hard \ac{sfmds} in terms of accuracy, which can be attributed to its softer constraints. Nevertheless, as shown in the attached video, qualitatively, both methods closely resemble the demonstrations. Regarding stability, results are reported using a total of $4064$ trajectories rolled out from a grid of initial conditions in $\mathrm{SE}(3)$. We note that a negligible portion of non-convergent trajectories arises due to the weak stability guarantees of the soft \ac{sfmds} ($<0.13\%$) and to edge cases of hard \ac{sfmds} ($<0.03\%$).
\begin{figure*}
    \centering
    \includegraphics[width=.3\linewidth, trim=15 40 5 30, clip]{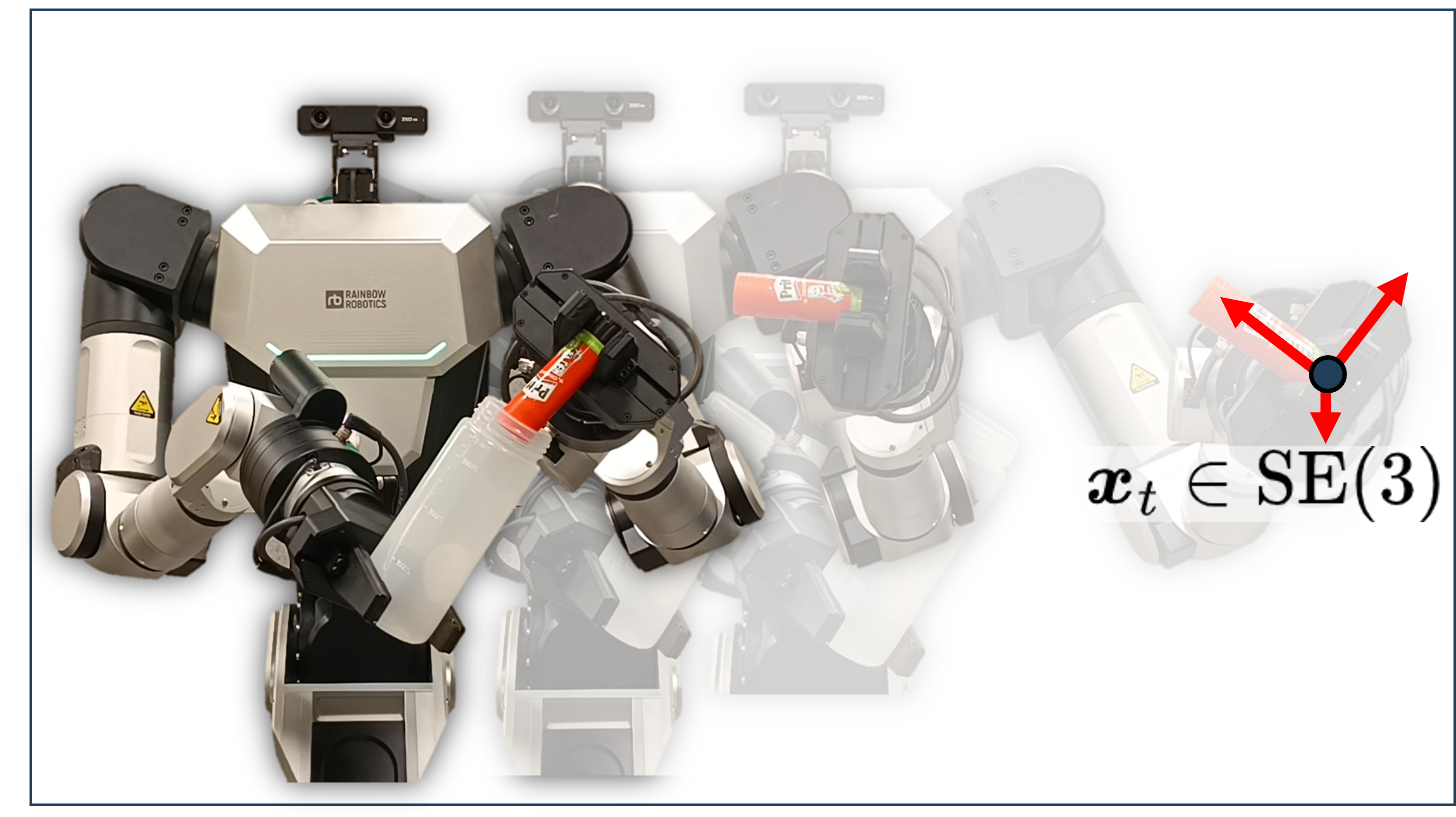}
    \includegraphics[width=.3\linewidth, trim=15 120 20 15, clip]{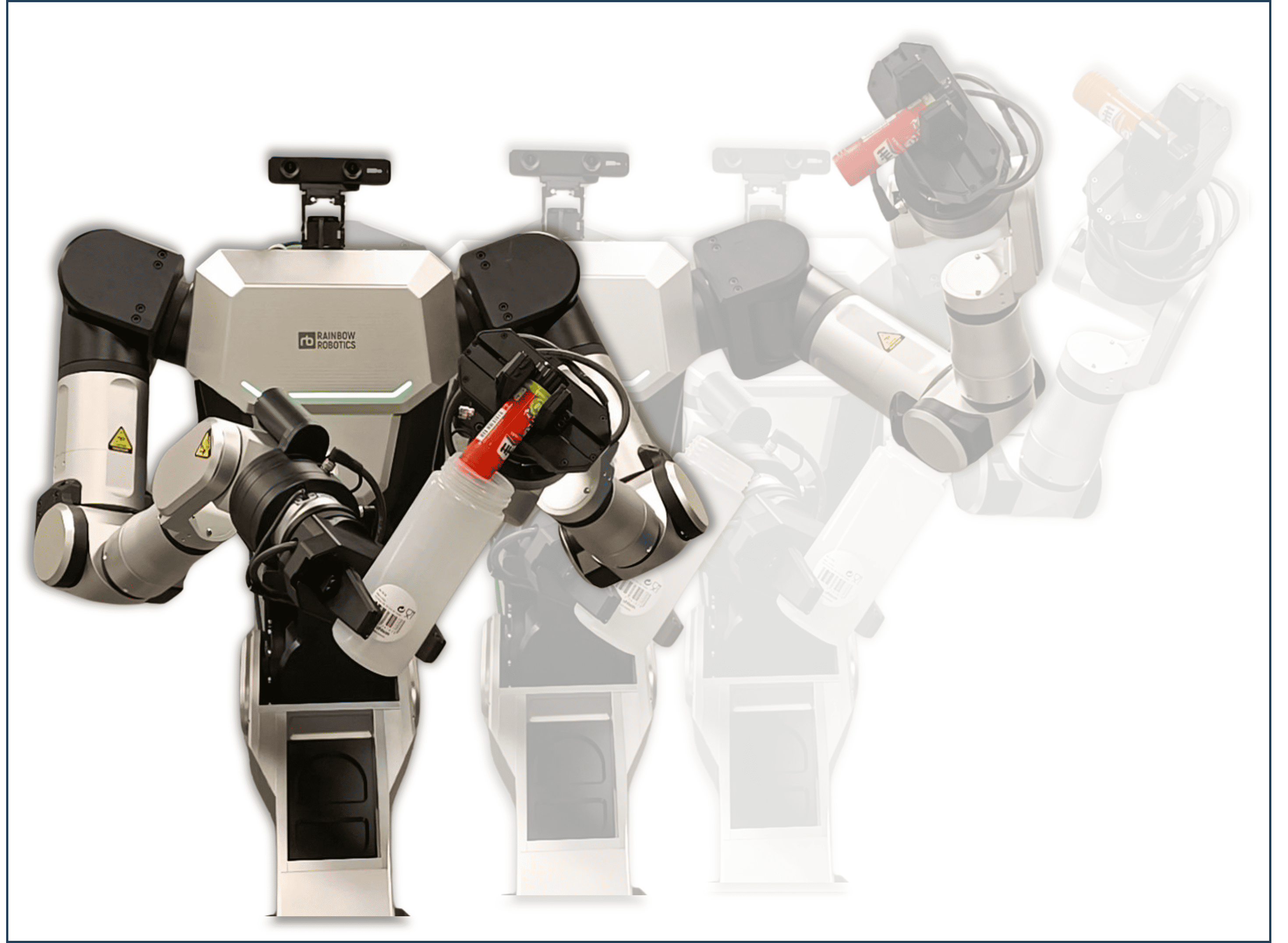}
    \includegraphics[width=.3\linewidth, trim=15 125 10 30, clip]{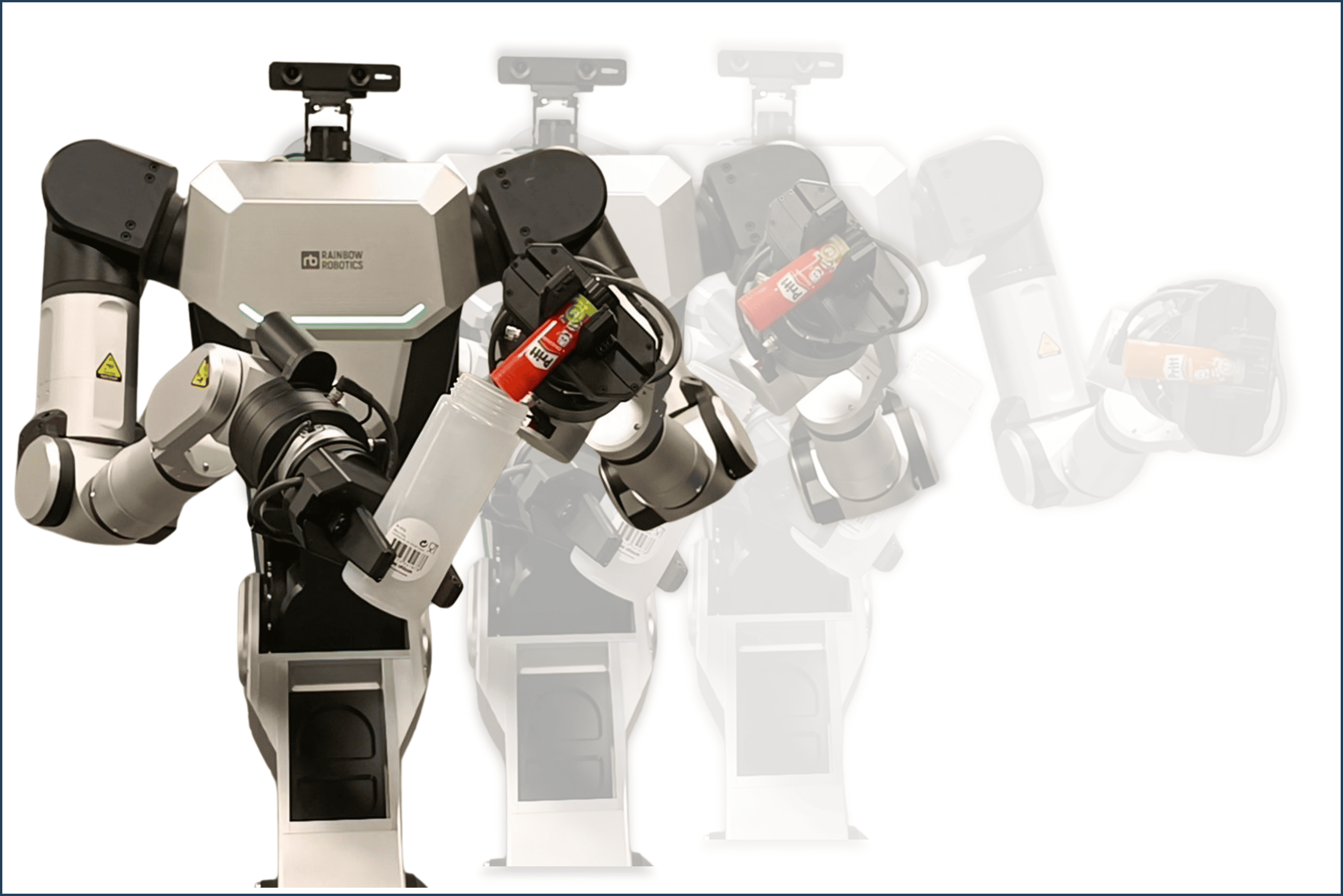}
    \caption{Sequences of robot poses along the hard \ac{sfmds} trajectories on the $\mathrm{SE}(3)$ insertion task for three different initial conditions.}
    \label{fig:robot_exps}
\end{figure*}
\subsection{Insertion Task on $\mathrm{SE}(3)$ with a Real Humanoid Robot}
\label{sec:robot_exp}
We validate \ac{sfmds} on a real robotic manipulation task, where a RB-Y1 humanoid robot inserts a cylindrical object into a container, see Fig.~\ref{fig:robot_exps}. The pose trajectory of the left end-effector is learned using \ac{sfmds} on $\mathrm{SE}(3)$ from $15$ demonstrations with different initial poses, collected using a master arm with the same kinematic structure as the robot. 
The robot’s end-effector is controlled via a Cartesian impedance controller provided by the manufacturer, running at $500$ Hz. At each control cycle, the desired velocity $\dot{\bm{x}}_t^{d}$, generated by the learned dynamics $f_{\theta}$, is integrated using a single Euler step to produce the pose reference for the controller. 

The frequency of the control loop is limited by the inference time of the learned model. Since evaluating flow-matching models requires an integration procedure, they are inherently slower than explicit models, making inference-time evaluation relevant. This situation is further exacerbated for hard \ac{sfmds} implemented via NODE, which relies on an additional inner integration loop at each inference step and the computation of the corresponding Jacobian. To alleviate this issue, we implemented a custom Taylor-series approximations of the Jacobian and its inverse, rather than relying on PyTorch's implementations. Although these approximations are less precise, they are significantly faster and were empirically found to be sufficiently accurate to achieve strong performance. 

We tested a learned hard \ac{sfmds} from both seen and unseen initial poses and introduced perturbations during execution. Three examples are provided in Fig.~\ref{fig:robot_exps} and detailed qualitative results are shown in the accompanying video. We observe that the learned model converges for most initial conditions, is robust to disturbances, and integrates seamlessly with standard robotic controllers in real time. A handful of failures occured due to workspace limits and self-collisions, which are not accounted for by \ac{sfmds}. 

\subsection{High-dimensional Reaction-diffusion System}
\label{subsec:exp_reactiondiffusion}
Finally, we evaluate the scalability of \ac{sfmds} on data generated from the evolution of a Gray-Scott reaction-diffusion (GSRD) system (see App.~\ref{app:exps_details} for details). This system produces two concentration fields $u$ and $v$ that can be visualized on a 2D grid. 
We represent both fields using $29\times29$ values, leading to a state of dimension $\dim(\mathcal{X})=2\times29\times29=1682$ (one field per channel).
We consider an instance of the GSRD system that showcases asymptotic stability (see in App.~\ref{app:exps_details}). We obtained a dataset of $10$ trajectories generated from different initial conditions, each consisting of $278$ frames. 
We learn a \ac{sfmds} with vector field $u_\theta$ parameterized via a UNet-like architecture to exploit the spatial correlations present in the GSRD system's concentration fields. For hard \ac{sfmds}, $\psi_\theta$ is parameterized by \emph{Glow} blocks~\cite{kingma2018glow}, which are invertible. We also consider behavior cloning (BC$_{\text{FMDS}}$) as a baseline. 
\input{Tables/high_dim_results}

\begin{figure*}
    \centering
    \subfloat[]{\includegraphics[width=1.0\linewidth]{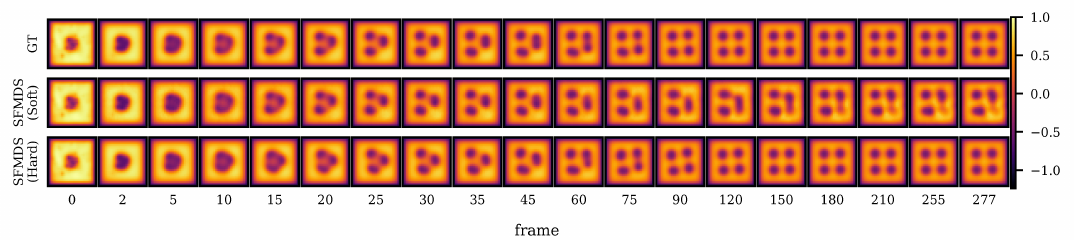}} \\
    \vspace{-10pt}
    \subfloat[]{\includegraphics[width=1.0\linewidth]{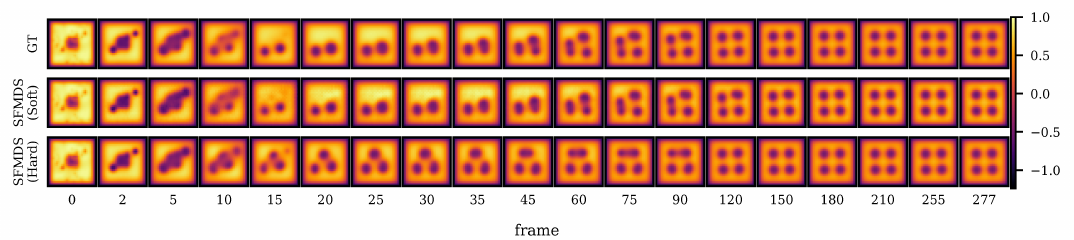}}
    \caption{Ground-truth (GT) and generated trajectories via Soft and Hard SFMDS associated with two different demonstrations ((a) and (b), respectively).}
    \label{fig:high_dim_stacked_rollouts}
\end{figure*}

Fig.~\ref{fig:high_dim_stacked_rollouts} shows two ground-truth trajectories for the concentration field $u$, and the corresponding trajectories generated via soft and hard \ac{sfmds}. 
In Fig~\ref{fig:high_dim_stacked_rollouts}(a), soft \ac{sfmds} converges to a spurious attractor. In contrast, hard \ac{sfmds} always converges to the equilibrium, even if the dynamical system temporarily diverges from the ground-truth, as in Fig~\ref{fig:high_dim_stacked_rollouts}(b).

\begin{table}[t]
\centering
\small
\caption{Avg. inference frequency (Hz) $\uparrow$ using an \emph{NVIDIA GeForce RTX 4090} GPU.} 
\label{tab:comp_cost}
\vspace{-0.15cm}
\begin{tabular*}{.8\linewidth}{l@{\extracolsep{\fill}}S[table-format=3.2]@{\extracolsep{\fill}}S[table-format=3.2]@{\extracolsep{\fill}}S[table-format=2.2]}
\toprule
Method & {$\mathbb{R}^2$} & {$\mathrm{SE}(3)$} & {$\mathbb{R}^{1682}$} \\
\midrule
BC$_{\text{FMDS}}$ / Soft \ac{sfmds} & 341.78 & 148.01 & 48.79 \\
Hard \ac{sfmds} & 17.14$^\dagger$ & 12.66$^\dagger$ & 14.03$^\ddagger$ \\
\bottomrule
\end{tabular*}
\vspace{2pt}
\footnotesize{\raggedright $\dagger$ $\psi_\theta$ parameterized using NODE. \hfill \\
$\ddagger$ $\psi_\theta$ parameterized via Glow blocks (see App~\ref{app:exps_details}). \hfill}
\end{table}
As in previous sections, we assess accuracy by generating rollouts of $f_{\theta}$ from the dataset's initial conditions. 
Table~\ref{tab:high_dim_metrics} reports the achieved performance assessed quantitatively via pixel-wise RMSE and Chamfer distances, showing that both \ac{sfmds} variants can accurately reproduce the demonstrated trajectories, outperforming BC$_{\text{FMDS}}$. 
To assess the stability of the learned dynamical systems, we generate rollouts starting from $1024$ out-of-distribution states (sampled from a uniform distribution $\text{Unif}([-1,1])$). Table~\ref{tab:high_dim_metrics} show that both \ac{sfmds} variants consistently reach the equilibrium, except for a few spurious attractors in soft \ac{sfmds}. 
Additional experiments and implementation details are reported in App.~\ref{app:exps_details}.

\subsection{Inference Times}
Table~\ref{tab:comp_cost} reports the inference frequency of the different models. Although the frequency decreases with the dimension, even hard \ac{sfmds} achieves control frequencies of approximately $13$ Hz, which is sufficient for a wide range of manipulation tasks.

%% file: Tables/unimodal_results.tex
\begin{table}
    \centering
    \scriptsize
    \caption{Accuracy and stability on the LASA and LASA $\torus^2$ datasets.}
    \vspace{-5pt}
    \begin{tabular*}{\linewidth}{@{}l@{\extracolsep{\fill}}l@{\extracolsep{\fill}}c@{\extracolsep{\fill}}c@{\extracolsep{\fill}}c@{\extracolsep{\fill}}c@{}}
        \toprule
        &\multirow{3}{*}{Method} &\multicolumn{3}{c}{Accuracy} &  Stability\\
        \cmidrule{3-5} \cmidrule{6-6} 
        & & RMSE$\downarrow$ & DTDW$\downarrow$ & FD$\downarrow$ & $\%$ Unsuc. Trajs. $\downarrow$\\ 
        \midrule
        \multirow{7}{*}{\rotatebox[origin=c]{90}{$\mathbb{R}^2$--SFMDS}} 
        & Hard$_{\text{RealNVP}}$ 
        & $\text{2.636}_{\pm \text{1.745}}$ 
        & $\text{1.151}_{\pm \text{0.933}}$ 
        & $\text{2.654}_{\pm \text{1.975}}$ 
        & $\text{0.000}$ \\
        & Hard$_{\text{1S RealNVP}}$ 
        & $\text{2.811}_{\pm \text{1.794}}$ 
        & $\text{1.139}_{\pm \text{0.864}}$ 
        & $\text{2.564}_{\pm \text{1.685}}$ 
        & $\text{0.044}$ \\
        & Hard$_\text{NODE}$  
        & $\text{2.661}_{\pm \text{1.867}}$ 
        & $\text{1.111}_{\pm \text{0.813}}$ 
        & $\text{2.622}_{\pm \text{1.837}}$ 
        & $\text{0.000}$ \\
        & Hard$_{\text{1S NODE}}$  
        & $\text{2.896}_{\pm \text{1.948}}$ 
        & $\text{1.106}_{\pm \text{0.816}}$ 
        & $\text{2.569}_{\pm \text{1.844}}$ 
        & $\text{0.000}$ \\
        & Soft            
        & $\text{2.400}_{\pm \text{1.585}}$ 
        & $\text{1.061}_{\pm \text{0.763}}$ 
        & $\text{2.547}_{\pm \text{1.695}}$ 
        & $\text{0.000}$ \\
        & Soft$_\text{1S}$             
        & $\text{2.471}_{\pm \text{1.589}}$ 
        & $\text{1.053}_{\pm \text{0.698}}$ 
        & $\text{2.520}_{\pm \text{1.611}}$ 
        & $\text{0.000}$ \\
        & BC$_{\text{FMDS}}$            
        & $\text{2.137}_{\pm \text{1.606}}$ 
        & $\text{0.961}_{\pm \text{0.862}}$ 
        & $\text{2.382}_{\pm \text{1.886}}$ 
        & $\text{40.163}$ \\

        \midrule 
        \multirow{4}{*}{\rotatebox[origin=c]{90}{$\mathbb{R}^2$--SoTA}} 
        & CLF-DM~\citep{khansari-zadeh2014learning}  
        & $\text{3.492}_{\pm \text{2.716}}$ 
        & $\text{1.901}_{\pm \text{2.547}}$ 
        & $\text{4.205}_{\pm \text{5.022}}$ 
        & 1.133 \\
        & Euc. Flows\citep{Rana2020:EuclideanizingFlows} 
        & $\text{2.926}_{\pm \text{2.415}}$ 
        & $\text{1.519}_{\pm \text{1.980}}$ 
        & $\text{3.278}_{\pm \text{3.401}}$ 
        & 2.144 \\
        & Euc. Flows$_\text{NODE}$  
        & $\text{2.731}_{\pm \text{2.220}}$ 
        & $\text{1.351}_{\pm \text{1.535}}$ 
        & $\text{3.184}_{\pm \text{3.214}}$ 
        & 0.755 \\
        & PUMA~\citep{perez2024puma}           
        & $\text{2.202}_{\pm \text{1.448}}$ 
        & $\text{0.963}_{\pm \text{0.597}}$ 
        & $\text{2.286}_{\pm \text{1.384}}$ 
        & 0.000 \\
        \midrule \midrule
        \multirow{4}{*}{\rotatebox[origin=c]{90}{\shortstack{$\torus^2$--\\SFMDS\\\mbox{}}}} 
        & Hard$_{\text{RealNVP}}$ 
        & $\text{0.124}_{\pm \text{0.080}}$ 
        & $\text{0.073}_{\pm \text{0.045}}$ 
        & $\text{0.168}_{\pm \text{0.118}}$ 
        & $\text{0.000}$ \\
        & Hard$_\text{NODE}$  
        & $\text{0.119}_{\pm \text{0.081}}$ 
        & $\text{0.077}_{\pm \text{0.075}}$ 
        & $\text{0.180}_{\pm \text{0.147}}$ 
        & $\text{0.000}$ \\
        & Soft            
        & $\text{0.118}_{\pm \text{0.091}}$ 
        & $\text{0.080}_{\pm \text{0.108}}$ 
        & $\text{0.193}_{\pm \text{0.235}}$ 
        & $\text{0.000}$ \\
        & BC$_{\text{FMDS}}$            
        & $\text{0.087}_{\pm \text{0.058}}$ 
        & $\text{0.056}_{\pm \text{0.036}}$ 
        & $\text{0.140}_{\pm \text{0.082}}$ 
        & $\text{32.248}$ \\
        \midrule
        \multirow[-6ex]{2}{*}{\rotatebox[origin=c]{90}{\shortstack{$\torus^2$--\\SoTA}}} 
        & LieImFlow~\citep{Urain22:StableSE3} 
        & $\text{0.134}_{\pm \text{0.116}}$ 
        & $\text{0.089}_{\pm \text{0.111}}$ 
        & $\text{0.205}_{\pm \text{0.241}}$ 
        & 0.026 \\
        & PUMA 
        & $\text{0.097}_{\pm \text{0.061}}$ 
        & $\text{0.061}_{\pm \text{0.035}}$ 
        & $\text{0.148}_{\pm \text{0.084}}$ 
        & 0.000 \\

        \bottomrule
    \end{tabular*}
    \vspace{2pt}

    \label{tab:lasa_ablation_results}
\end{table}

%% file: Tables/multimodal_results.tex
\begin{table}
    \centering
    \scriptsize
    \caption{Accuracy and stability on multimodal datasets in $\mathbb{R}^2$ and $\mathrm{SE}(3)$}
    \vspace{-5pt}
    \begin{tabular*}{\linewidth}{@{}l@{\extracolsep{\fill}}l@{\extracolsep{\fill}}c@{\extracolsep{\fill}}c@{\extracolsep{\fill}}c@{}}
        \toprule
        &\multirow{4}{*}{Method} &\multicolumn{2}{c}{Accuracy} & Stability \\
        \cmidrule{3-4} \cmidrule{5-5} 
        & & \multicolumn{2}{c}{Chamfer dist.} & \multirow{3}{*}{$\%$ Unsuc. Trajs. $\downarrow$}\\
        & & $\text{CH}(G, D)\downarrow$ & $\text{CH}(D, G)\downarrow$ & \\ 
        & & \hfill$^{\times10^{-3}}$ & \hfill$^{\times10^{-3}}$ & \hfill\\ [-5pt]
        \midrule
        \multirow{3}{*}{\rotatebox[origin=c]{90}{\shortstack{$\mathbb{R}^2$--\\SFMDS}}}
            & Hard$_\text{NODE}$  
            & $\text{3.828}_{\pm \text{5.178}}$ 
            & $\text{3.724}_{\pm \text{3.613}}$ 
            & $\text{0.000}$ \\
            & Soft                
            & $\text{1.402}_{\pm \text{0.783}}$ 
            & $\text{2.512}_{\pm \text{0.991}}$ 
            & $\text{0.000}$ \\
            & BC$_{\text{FMDS}}$                  
            & $\text{3.271}_{\pm \text{5.707}}$ 
            & $\text{2.501}_{\pm \text{1.138}}$ 
            & $\text{27.978}$ \\
        \midrule
        \multirow{3}{*}{\rotatebox[origin=c]{90}{\shortstack{$\mathbb{R}^2$--\\SoTA}}} 
            & Euc. Flows               
            & $\text{2.530}_{\pm \text{0.831}}$ 
            & $\text{94.13}_{\pm \text{60.88}}$ 
            & 23.056 \\
            & Euc. Flows$_\text{NODE}$ 
            & $\text{3.392}_{\pm \text{1.517}}$ 
            & $\text{36.10}_{\pm \text{21.38}}$ 
            & 0.089 \\
            & PUMA                     
            & $\text{4.018}_{\pm \text{2.604}}$ 
            & $\text{51.69}_{\pm \text{38.88}}$ 
            & 0.000 \\
        \midrule \midrule
        
        \multirow{3}{*}{\rotatebox[origin=c]{90}{\shortstack{$\mathrm{SE}(3)$--\\SFMDS}}}

        & Hard$_\text{NODE}$  
        & $\text{13.152}_{\pm \text{2.079}}$ 
        & $\text{14.783}_{\pm \text{2.215}}$ 
        & $\text{0.000}$ \\
        & Soft                
        & $\text{7.896}_{\pm \text{1.418}}$ 
        & $\text{10.447}_{\pm \text{1.505}}$ 
        & $\text{0.038}$ \\
        & BC$_{\text{FMDS}}$                  
        & $\text{8.191}_{\pm \text{1.940}}$ 
        & $\text{10.548}_{\pm \text{1.845}}$ 
        & $\text{21.355}$ \\
        \bottomrule
    \end{tabular*}
    \label{tab:multimodal_ablation_results}

\end{table}

%% file: Tables/high_dim_results.tex
\begin{table}
    \centering
    \scriptsize
    \caption{Accuracy and stability on the GSRD system.}
    \vspace{-5pt}
    \begin{tabular*}{\linewidth}{@{}l@{\extracolsep{\fill}}c@{\extracolsep{\fill}}c@{\extracolsep{\fill}}c@{\extracolsep{\fill}}c@{}}
        \toprule
       \multirow{4}{*}{Method} &\multicolumn{3}{c}{Accuracy} &  Stability\\
        \cmidrule{2-4} \cmidrule{5-5} 
         &\multirow{2}{*}{RMSE$\downarrow$} & \multicolumn{2}{c}{Chamfer dist.} & \multirow{2}{*}{$\%$ Unsuc. Trajs. $\downarrow$}\\
        &   & $\text{CH}(G, D)$$\downarrow$& $\text{CH}(D, G)$$\downarrow$\\ \midrule
        SFMDS$_{\text{Hard}}$ 
        & $\text{0.091}_{\pm \text{0.033}}$ 
        & $\text{1.061}_{\pm \text{1.744}}$ 
        & $\text{1.330}_{\pm \text{1.756}}$ 
        & $\text{0.000}$\\
        SFMDS$_{\text{Soft}}$ 
        & $\text{0.094}_{\pm \text{0.024}}$ 
        & $\text{2.607}_{\pm \text{0.504}}$ 
        & $\text{3.238}_{\pm \text{0.905}}$ 
        & $\text{2.344}$\\
        BC$_{\text{FMDS-clipped}}$ 
        & $\text{0.600}_{\pm \text{0.006}}$ 
        & $\text{7.404}_{\pm \text{1.896}}$ 
        & $\text{21.036}_{\pm \text{12.291}}$ 
        & $\text{100.0}$\\
        BC$_{\text{FMDS}}$           
        & $\text{4.905}_{\pm \text{0.197}}$ 
        & $\text{8.691}_{\pm \text{2.077}}$ 
        & $\text{136.173}_{\pm \text{147.995}}$ 
        & $\text{100.0}$\\
        \bottomrule
    \end{tabular*}
  
    \label{tab:high_dim_metrics}
\end{table}

%% file: Sections/10_Conclusions.tex
\section{Conclusions}
This paper introduced \emph{Stable Flow Matching Dynamical Systems} (SFMDS), a dynamical-system-based imitation learning framework that combines the generative power of flow matching with the stability guarantees of dynamical-system-based policies. SFMDS models the velocity at each state through a flow-matching process while restricting the set of admissible solutions using control-theoretic constraints, thus extending concepts from latent Lyapunov stability analysis to flow matching dynamical system. We proposed two \ac{sfmds} variants encoding stability either via soft penalty terms or via hard constraints. We validated their performance through extensive experiments on benchmark datasets, simulations and real robotic tasks. Finally, we demonstrated scalability to a $1682$-dimensional dynamical system.

Several research directions could further extend the capabilities of the \ac{sfmds} framework. First, many robotic tasks cannot be adequately described by a single goal pose: Extending the framework to handle convergence to sets of goal states would therefore increase its applicability. Second, incorporating the robot's physical constraints directly into the dynamical system driving the end-effector motion would improve robustness, as most failures in the real-robot experiments were caused by constraint violations such as imminent self-collisions. Finally, richer policy conditioning, for example, through images or point clouds, could broaden the applicability of \ac{sfmds} and motivate the study of stability in these settings.

%% file: Sections/09_Appendix/Appendix.tex
\appendices
\input{Sections/09_Appendix/Method_Details}
\input{Sections/09_Appendix/Highdim_System}

%% file: Sections/09_Appendix/Method_Details.tex
\input{Sections/09_Appendix/Edge_Cases_discrete_u}
\input{Sections/09_Appendix/Lie_Groups}

%% file: Sections/09_Appendix/Edge_Cases_Discrete_u.tex
\section{Edge Cases for Hard-Constrained Methods}
\label{app:edge_cases_u}

As discussed in Secs.~\ref{subsec:discrete-time-systems} and~\ref{subsec:exp_benchmarks}, a few rare edge cases compromising the stability of hard \ac{sfmds} might arise from the discretization of the dynamics $u_{\theta}$.
In this appendix, we describe how these cases can be addressed to increase the robustness of hard \ac{sfmds}.

\subsubsection{Escaping $\dot{\mathcal{X}}_{\mathrm{A}}$}
\begin{figure}[t]
    \centering
    \begin{overpic}[width=\linewidth]{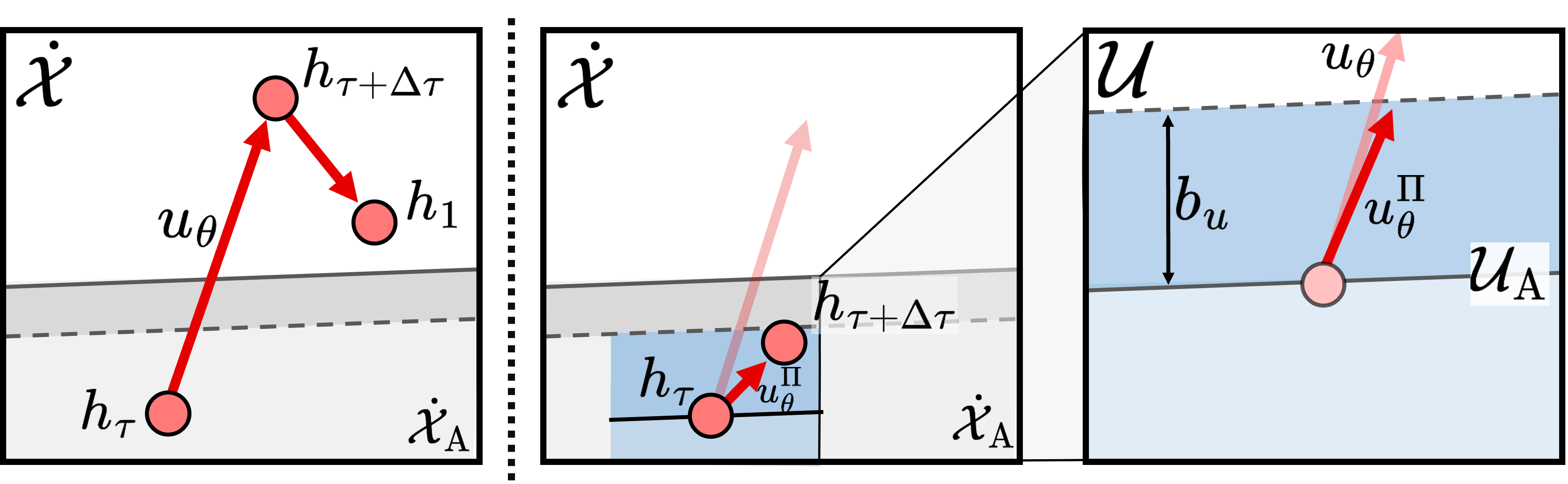}
        \put(13,-5){\footnotesize (a)}
        \put(47,-5){\footnotesize (b)}
        \put(81,-5){\footnotesize (c)}
    \end{overpic}
    \vspace{0.05cm}
    \caption{Failure edge case caused by discretizing $u_\theta$. (a) With one integration step remaining, the state $\bm{h}_\tau$ jumps outside $\dXa$ and fails to converge back after the final step. (b) Constraining the vector prevents $\bm{h}_\tau$ from leaving $\dXa$. (c) Detailed view of the admissible set $\dUa$ for this case.}
    \label{fig:edge_case_u}
\end{figure}

Next, we consider the case when $\dXa$ is a half-space. However, the same approach can be employed in the case when $\dXa$ is a ball. When $\bm{h}_{\tau} \!\in\! \dXa$, the ReLU-based projection in~\eqref{eq:proj-vector} leaves the vector field $u_\theta$ unconstrained. In such cases, if $u_\theta$ induces motion toward the boundary of the admissible half-space, discretization effects may cause the state $\bm{h}_{\tau}$ to step outside $\dXa$ in a single integration step, as shown in Fig.~\ref{fig:edge_case_u}(a). Although the constrained $u_\theta$ will subsequently enforce convergence back toward $\dXa$, this may be insufficient when only a limited number of integration steps remain. As a result, the trajectory may fail to converge to $\dXa$ by $\tau = 1$. Alternatively, it can enter oscillatory regimes near the boundary, effectively making set membership dependent on the particular phase of the oscillation at which the integration terminates.

To address this issue, whenever $u_{\theta}$ induces motion that drives $\bm{h}_{\tau}$ toward the boundary of $\dXa$, we constrain the vector field such that a single Euler integration step cannot cause the state to escape $\dXa$. To this end, in these situations, we remove the activation $\sigma_r$ from~\eqref{eq:proj-vector}, yielding
\begin{equation}
\label{eq:proj-vector-hard}
\delta_{\bm{h}} = -\frac{\bm{n}_{\bm{h}}^{\trsp}\bm{h}_{\tau} + m_{\bm{h}}}{\|\bm{n}_{\bm{h}}\|^{2}} \, \bm{n}_{\bm{h}}.
\end{equation}
We then use $\delta_{\bm{h}}$ to compute the offset $b_u$, which in this case is treated as a \emph{maximum} allowable offset toward the boundary, satisfying
\begin{equation}
\label{eq:bu_max_app}
    b_u = \lambda_{\bm{h}} \, \lVert \delta_{\bm{h}} \rVert.
\end{equation}
This bound is subsequently used to define the margin $m_u$, which enforces the condition
\begin{equation}
\label{eq:def-mB_app_hard}
   -\delta_{\bm{h}}^\trsp u_{\theta} - m_u < 0, \quad \forall \bm{h}_{\tau} \in \dXa.
\end{equation}
Since the margin $m_u$ is now subtracted, the half-space offset is applied in the opposite direction compared to~\eqref{eq:def-mB}, effectively preventing outward motion across the boundary (see Fig.~\ref{fig:edge_case_u}(b)-(c).

\subsubsection{Overfitting of $\psi_{\theta}$}
\label{sec:psi_overfitting}
In cases where the transformation $\psi_{\theta}$ varies too sharply, for instance, due to overfitting, the discretized dynamics may violate the stability conditions even when the admissible set is defined as a ball in the latent space. In this case, the latent dynamics may remain stable, while the corresponding deformation induced by $\bm{J}_{\psi_{\theta}}^{-1}$ changes too abruptly in task space. As a result, a single integration step may move the task-space trajectory in a manner inconsistent with the intended stability guarantees.
This issue can be mitigated in several ways. A first possibility is to regularize $\psi_{\theta}$ to promote smoothness, for example, by penalizing large Jacobian norms. Another option is to use invertible architectures that are empirically less prone to such sharp local deformations; for instance, NODE-based transformations are preferable to RealNVP-type mappings in this regard. The effect of discretization can also be reduced by using higher-order integration schemes or by decreasing the integration step size.
Finally, a promising alternative is to perform numerical integration directly in the latent space, where stability conditions are enforced by construction, and then map the integrated state back to task space through $\psi_{\theta}^{-1}$. We leave a systematic investigation of this direction for future work.

%% file: Sections/09_Appendix/Lie_Groups.tex
\section{Lie Groups: Exponential, logarithmic, and adjoint maps}
\label{app:lie_groups}

To construct and simulate dynamics on Lie groups, we use three operations: (1) in hard \ac{sfmds}, we employ the logarithmic map to represent states in the group's Lie algebra; (2) to simulate the dynamics, we use a forward Euler scheme adapted to Lie groups, where one integration step is given by
\begin{equation}
\label{eq:lie_group_euler}
    \bm{x}_{t+\Delta t} = \bm{x}_t \exp\bigl(\Delta t\, \mathrm{Ad}_{\bm{x}}^{-1} f_\theta(\bm{x}_t, \bm{\omega}_t)\bigr),
\end{equation}
where $\Delta t$ is the integration step size, and (3) we use the adjoint operator to map vectors from local frames to global frames at the identity, as shown in~\eqref{eq:u_lie_integration}. Note that, in hard \ac{sfmds}, the input to $f_\theta$ is the Lie algebra representation $\bm{\xi}_t = \log(\bm{x}_t)$. Thus, strictly speaking, the vector field should be written as $f_\theta(\bm{\xi}_t, \bm{\omega}_t)$ in~\eqref{eq:lie_group_euler}. To use a unified notation for both soft and hard \ac{sfmds}, we assume that, in the hard case, the logarithmic map is implemented as the first frozen layer of $f_\theta$. This allows us to write the vector field consistently as $f_\theta(\bm{x}_t, \bm{\omega}_t)$.
Next, we detail these operations for the specific Lie groups considered in this work.

\subsubsection{Torus $\mathbb{T}^2$}
The $2$-torus is identified as the Lie group ${\mathbb{T}^2 = \sS^1 \times \sS^1}$. Since $\sS^1$ is isomorphic to the set of unit complex numbers $\{e^{i\theta} \mid \theta \in \mathbb{R}\} \subset \mathbb{C}$, elements on the torus can be written as $\bm{x}_t = (e^{i\theta_{1, t}}, e^{i\theta_{2, t}}) \in \mathbb{T}^2 \subset \mathbb{C}^2$~\cite[Ch.~7]{Lee13:SmoothManifolds}. Its Lie algebra is $\mathfrak{t}^2 \cong \mathbb{R}^2$, whose elements can be written as $\bm{\xi}_t=(\theta_{1, t}, \theta_{2, t})$. 
The exponential and logarithmic maps are then constructed as,
\begin{align}
\exp_{\mathbb{T}^2}(\bm{\xi}_t) &= (e^{i\theta_{1,t}}, e^{i \theta_{2,t}}) \in \mathbb{C}^2, \label{eq:exp_map_T2} \\
\log_{\mathbb{T}^2}(\bm{x}_t) &= \big (\mathrm{wrap}( \theta_{1,t}), \mathrm{wrap}( \theta_{2,t}) \big) \in \mathbb{R}^2, \label{eq:log_map_T2}
\end{align}
where the $\mathrm{wrap}$ operator maps an angle to its principal value in the interval $[-\pi, \pi]$.
Since the 2-torus group is Abelian, the adjoint matrix is trivial, leading to
\begin{equation}
    \mathrm{Ad}_{\bm{x}} = I_2, \quad \forall \bm{x}_t \in \mathbb{T}^2.
\end{equation}

\subsubsection{Special Euclidean group $\mathrm{SE}(3)$}
Each element of $\mathrm{SE}(3)$ is a pose 
\begin{equation*}
\bm{x}_t = \begin{bmatrix}
\bm{R}_t & \bm{p}_t \\
0 & 1
\end{bmatrix} \in \mathrm{SE}(3),
\end{equation*}
where $\bm{R}_t \in \mathrm{SO}(3)$ and $\bm{p}_t \in \mathbb{R}^3$ denote the rotation and translation components, respectively.
A Lie algebra element is written as
\begin{equation}
\bm{\xi}_t =
\begin{bmatrix}
[\bm{\omega}_t]_\times & \bm{v}_t \\
\bm{0}^\top & 0
\end{bmatrix}
\in \mathfrak{se}(3).
\end{equation}
The exponential map is defined as~\cite{sola2018micro},
\begin{equation}
\exp_{\mathrm{SE}(3)}({\bm{\xi}}_t) =
\begin{bmatrix}
\exp_{\mathrm{SO}(3)}([\bm{\omega}_t]_\times) & \bm{J}(\bm{\omega}_t)\,\bm{v}_t \\
\bm{0}^\top & 1
\end{bmatrix},
\end{equation}
where the rotational exponential is given by Rodrigues' formula~\cite{sola2018micro},
\begin{equation}
\begin{aligned}
\exp_{\mathrm{SO}(3)}([\bm{\omega}_t]_\times)
&= \bm{I}
+ \frac{\sin\theta_t}{\theta_t}[\bm{\omega}_t]_\times
+ \frac{1 - \cos\theta_t}{\theta_t^2}[\bm{\omega}_t]_\times^2,
\\  \theta_t &= \|\bm{\omega}_t\|,
\end{aligned}
\end{equation}
and the left Jacobian $\bm{J}(\bm{\omega}_t)$ is
\begin{equation}
\bm{J}(\bm{\omega}_t) =
\bm{I}
+ \frac{1 - \cos\theta_t}{\theta_t^2}[\bm{\omega}_t]_\times
+ \frac{\theta_t - \sin\theta_t}{\theta_t^3}[\bm{\omega}_t]_\times^2.
\end{equation}
The logarithm is obtained by first computing
\begin{align}
[\bm{\omega}]_\times &= \log_{\mathrm{SO}(3)}(\bm{R}_t)
= \frac{\theta_t}{2\sin\theta_t}(\bm{R}_t - \bm{R}_t^\top),
\\
\theta_t &= \cos^{-1}\!\left(\frac{\mathrm{tr}(\bm{R}_t) - 1}{2}\right).
\end{align}
Then, the translational component is recovered using the inverse left Jacobian,
\begin{equation}
\bm{v}_t = \bm{J}(\bm{\omega}_t)^{-1} \bm{p}_t.
\end{equation}

Thus,
\begin{equation}
\log_{\mathrm{SE}(3)}(\bm{x}_t) =
\begin{bmatrix}
[\bm{\omega}_t]_\times & \bm{v}_t \\
\bm{0}^\top & 0
\end{bmatrix}.
\end{equation}

The adjoint matrix $\mathrm{Ad}_{\bm{x}} : \mathfrak{se}(3) \to \mathfrak{se}(3)$ has the form~\cite{sola2018micro},
\begin{equation}
 \mathrm{Ad}_{\bm{x}} = \begin{bmatrix}
\bm{R}_t & 0 \\
[\bm{p}_t]_\times \bm{R}_t & \bm{R}_t
\end{bmatrix},
\end{equation}
where $[\bm{p}_t]_\times$ denotes the skew-symmetric matrix associated with the cross product.

%% file: Sections/09_Appendix/Highdim_System.tex
\begin{figure*}
    \centering
    \includegraphics[width=\linewidth]{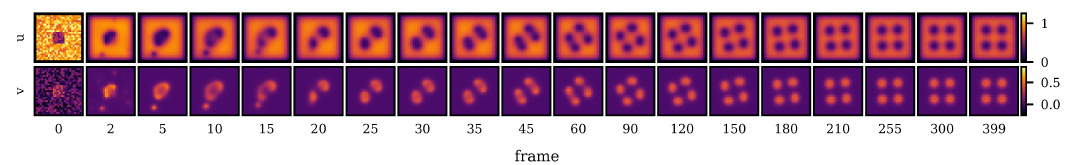}
    \vspace{-20pt}
    \caption{Example of ground-truth trajectory for the $u$ and $v$ fields. Note that, at each frame, both fields are correlated although their numerical values differ.}
    \label{fig:gt_rd_trajectory_example_fields}
\end{figure*}

\begin{figure*}
    \centering
    \includegraphics[width=.9\linewidth]{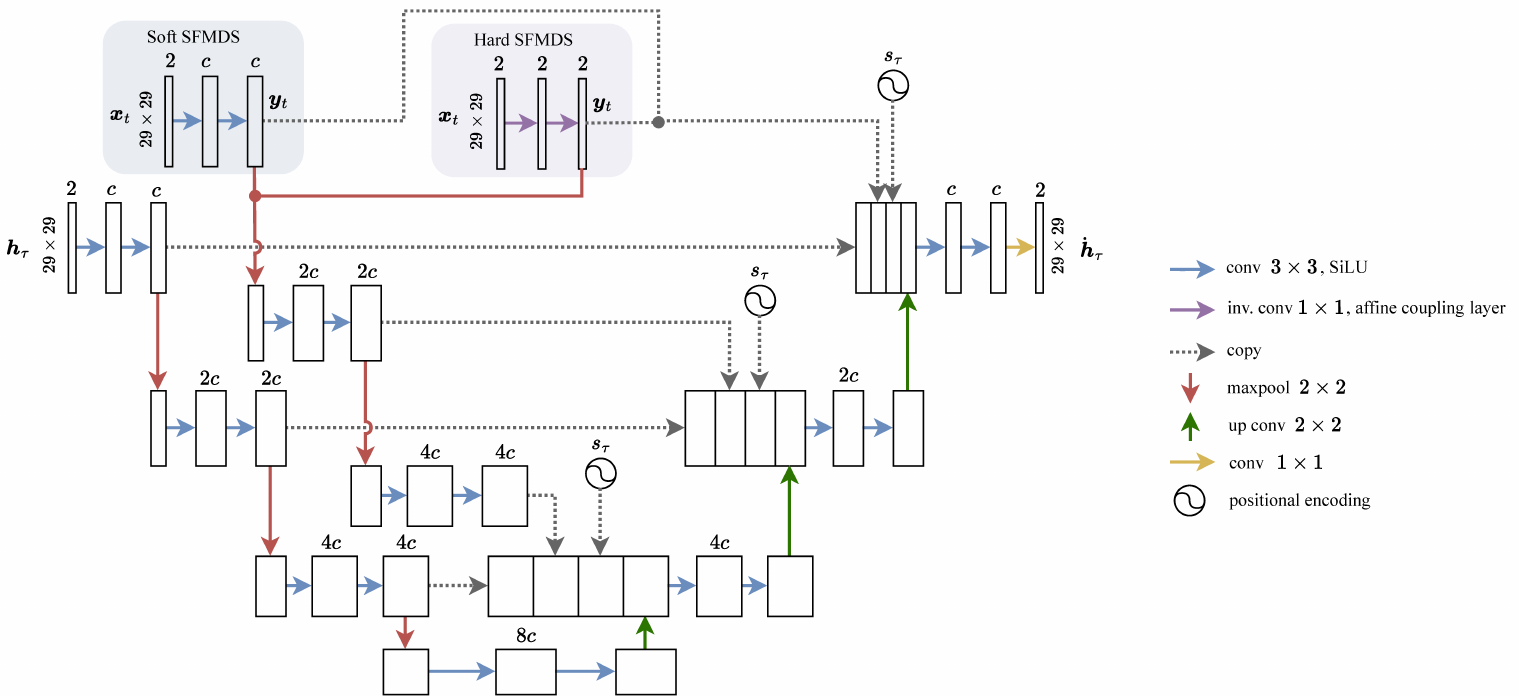}
    \caption{Diagram of the UNet-like architecture used to parameterize $u_\theta$, for both soft and hard \ac{sfmds}.}
    \label{fig:custom_unet}
\end{figure*}

\section{High-dimensional Reaction Diffusion System}
\label{app:exps_details}
We consider the Gray-Scott reaction-diffusion system, which models the chemical reaction given by
\begin{align*}
    U+2V & \rightarrow 3V\\
    V &\rightarrow P
\end{align*}
where $U$ is consumed, $V$ produced, and $P$ is an inert product.
This system is governed by the system of ODE
\begin{align}
    \frac{\partial u}{\partial t} = D_u\nabla^2u - uv^2 + F(1-u) \label{eq:gs_rd_du},\\
    \frac{\partial v}{\partial t} = D_v\nabla^2v + uv^2 - (F+k)v \label{eq:gs_rd_dv},
\end{align}
where $u$ and $v$ are the concentrations of $U$ and $V$, $D_u$ and $D_v$ are their diffusion coefficients, and the parameters $F$ and $k$ control the rates at which $U$ is fed to the system, and $V$ is removed from it, respectively.
This reaction-diffusion system has been studied extensively due to the complex spatiotemporal patterns it can exhibit \citep{pearson1993complex}.
In this experiment, we focus on a particular instance of this system, whose solution is characterized by spots that emerge after a perturbation and multiply until filling the available space\footnote{Referred to as ``pattern $\lambda$'' by \cite{pearson1993complex}.}. This is accomplished by setting $D_u=0.16, D_v=0.06, F=0.034,$ and $k=0.065$. Moreover, enforcing Neumann boundary conditions and limiting the physical space in which the system evolves makes it reach the same steady state for different initial perturbations, which renders it a suitable testbed for assessing the stability of learned dynamical systems.


\begin{figure*}[h]
    \centering
    \includegraphics[width=\linewidth]{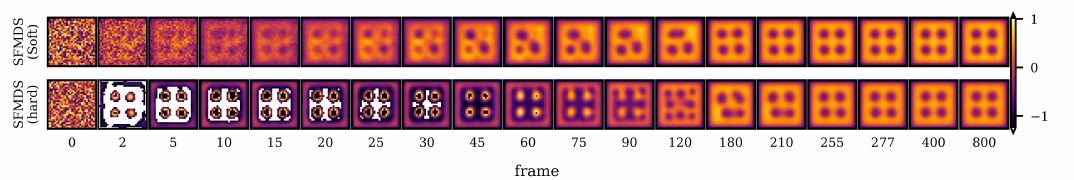}
    \caption{Examples of trajectories generated by Soft and Hard SFMDS when starting from OOD initial states. Out-of-bound pixels are white if below the lower bound, and black if above the upper bound.}
    \label{fig:high-dim-ood}
\end{figure*}

As in \cite{pearson1993complex}, we simulate the system via forward Euler integration. The spatial 2D grid is set to $29\times29$ points, and the initial state is set to $(u,v)=(1.0, 0.0)$ with a perturbation set to $u=0.5$, $ v=0.25$ in a square region of size $r=7$ in the center of the grid. Each point in the grid is then further perturbed by adding to it a uniform noise sample, $\epsilon\sim \text{Unif}([-s, s])$, where $s=0.25$. Fig.~\ref{fig:gt_rd_trajectory_example_fields} show examples of trajectories obtained when simulating the reaction-diffusion system with the conditions described above, recording frames every $10$ Euler integration steps. Note that the frames are unevenly spaced to highlight the fields' evolution, which tends to be more noticeable at the beginning of the trajectories.

To learn a model of this dynamical system, we construct a dataset by simulating the reaction-diffusion system and using the resulting trajectories, discarding the initial state frame. Although different trajectories may reach the steady state at a different frame number, we empirically found that simulating the reaction-diffusion system for $500$ frames suffices to reach the fixed point consistently. We then stablish a new maximum frame number for all trajectories by retrieving the frame index at which we observe no noticeable changes in the states' evolution. This results in a dataset of $10$ trajectories, each consisting of $278$ frames.  
We normalize the collected data via an independent min-max normalization of the concentration fields $u$ and $v$ to the interval $[-1, 1]$, using the numerical ranges that these fields take for all the simulated trajectories. The normalized fields are then stacked, making every data point in the dataset a $2\times29\times29$ image, i.e., $\dim(\dot{\mathcal{X}})=1682$.

We exploit the spatial correlation of the concentration fields by parameterizing the vector field $u_\theta$ of soft and hard \ac{sfmds} with the UNet-like neural architecture~\citep{ronneberger2015u} displayed in Fig.~\ref{fig:custom_unet}. 
The positional encoding module is implemented as described in~\cite{vaswani2017attention}, and its outputs are repeated and reshaped to match the expected dimension that allows its concatenation with the corresponding $\bm{h}_\tau$ and $\bm{x}_t$ embeddings at each skip connection. The hard SFMDS sub-neural network $\psi_\theta$ is parameterized by two \emph{Glow} blocks~\cite{kingma2018glow} without \emph{actnorm} layers. For all experiments, the parameter $c$ is set to $32$. To compute the auxiliary loss functions of soft \ac{sfmds}, we draw (ambient space) samples from a clipped standard normal distribution, where clipping constrains the numerical values of the samples to lie within $[-1,1]$. These samples contain values at the ambient space boundary since, as previously stated, a min-max normalization is performed over the reaction-diffusion trajectories used to construct the database. In addition, for the soft \ac{sfmds} variant, both the positively invariant set and asymptotic stability loss functions are computed performing pixel-wise projections, which we found to work better empirically.
%
%
%
The model rollouts are constructed by conditioning the learned vector field $u_\theta$ on ground-truth initial states for the first velocity prediction only. This velocity prediction is then used to predict the next state, which is then used as the new conditioning state for the following prediction. 





Additional examples of trajectories generated by the two \ac{sfmds} variants, starting from OOD initial states are displayed in Fig.~\ref{fig:high-dim-ood}, where out of bounds pixels are represented in different colors to improve visualization.
We observe that the behavior of the soft \ac{sfmds} variant is consistent with the positively invariant set constraint, as pixels remain bounded for the trajectory. In contrast, the hard \ac{sfmds} variant can generate states with pixels going out of bounds, since a positively invariant set is not enforced nor required in this case. Note that this is consistent with observations made for hard \ac{sfmds} in the benchmark datasets of Sec.~\ref{subsubsec:lasa_exps_main}. Finally, we note that, although the evolution of pixels during the generation of trajectories is not numerically bounded for the \ac{sfmds}, the steady state is consistently reached by both \ac{sfmds} variants given a sufficient prediction horizon. 